\documentclass[10pt]{article} % For LaTeX2e
\usepackage[preprint]{tmlr}

\usepackage{amsmath,amsfonts,bm}
\usepackage{wrapfig,lipsum}
\usepackage{natbib}
\usepackage{caption}
\usepackage[utf8]{inputenc}
\usepackage[english]{babel}
\usepackage{amssymb,amsmath,amsthm}

\usepackage{xcolor}
\definecolor{prune}{rgb}{0.44, 0.11, 0.11}
\definecolor{myblue}{rgb}{0, .5, 1}
\definecolor{maroon}{rgb}{0.5450, 0, 0}
\definecolor{darkred}{rgb}{0.5450, 0, 0}
\definecolor{RoyalBlue}{RGB}{0,100,170}
\definecolor{DarkBlue}{RGB}{20,70,200}
\definecolor{peach}{rgb}{1, 0.56, 0.56}
\definecolor{NotionGreen}{RGB}{15,123,108}
\definecolor{NotionOrange}{RGB}{217,115,13}
\definecolor{NotionRed}{RGB}{224,62,62}
\definecolor{red}{RGB}{224,62,62}
\definecolor{midgray}{RGB}{150,150,150}
\definecolor{lavender}{rgb}{0.75, 0.58, 0.89}
\definecolor{Indigo7}{RGB}{66, 99, 235}
\definecolor{Green7}{RGB}{55, 178, 77}
\definecolor{Yellow7}{RGB}{245, 159, 0}
\definecolor{Red7}{RGB}{240, 62, 62}

\newcommand{\frameworkname}{Chimera}
\newcommand{\nameframework}{Chimera}

\def\eqref#1{equation~\ref{#1}}
\def\1{\bm{1}}

\def\vh{{\bm{h}}}

\def\vu{{\bm{u}}}
\def\vv{{\bm{v}}}

\def\vx{{\bm{x}}}
\def\vy{{\bm{y}}}

\DeclareMathAlphabet{\mathsfit}{\encodingdefault}{\sfdefault}{m}{sl}
\SetMathAlphabet{\mathsfit}{bold}{\encodingdefault}{\sfdefault}{bx}{n}

\def\bA{{\bm{A}}}
\def\bB{{\bm{B}}}
\def\bC{{\bm{C}}}

\def\bI{{\bm{I}}}

\def\bL{{\bm{L}}}
\def\bM{{\bm{M}}}

\def\bV{{\bm{V}}}

\def\bX{{\bm{X}}}
\def\bY{{\bm{Y}}}

\def\gE{{\mathcal{E}}}

\def\gG{{\mathcal{G}}}

\def\gN{{\mathcal{N}}}

\def\gV{{\mathcal{V}}}
\def\gW{{\mathcal{W}}}
\def\gX{{\mathcal{X}}}

\def\sN{{\mathbb{N}}}

\theoremstyle{plain}
\newtheorem{theorem}{Theorem}[section]
\newtheorem{proposition}[theorem]{Proposition}

\newtheorem{definition}[theorem]{Definition}

\theoremstyle{remark}

\usepackage{hyperref}
\usepackage{url}
\usepackage{graphicx}
\usepackage{subcaption}
\usepackage{amsmath}
\usepackage{booktabs}

\usepackage[utf8]{inputenc} % allow utf-8 input
\usepackage[T1]{fontenc}    % use 8-bit T1 fonts
\usepackage{hyperref}       % hyperlinks
\usepackage{url}            % simple URL typesetting
\usepackage{booktabs}       % professional-quality tables
\usepackage{amsfonts}       % blackboard math symbols
\usepackage{nicefrac}       % compact symbols for 1/2, etc.
\usepackage{microtype}      % microtypography
\usepackage{bm, bbm}

\usepackage{microtype}
\usepackage{graphicx}

\usepackage{natbib}
\usepackage{amsmath}
\usepackage{amssymb}
\usepackage{mathtools}
\usepackage{amsthm}

\usepackage{amsfonts}       % blackboard math symbols
\usepackage{caption}
\usepackage{subcaption}
\usepackage{dsfont}
\usepackage{makecell}
\usepackage{multirow}
\usepackage{xfrac}
\usepackage{colortbl}
\usepackage{wrapfig,lipsum}
\usepackage{float}
\usepackage{pythonhighlight}
\usepackage{pgfplots}
\usepackage{bbm}
\usepackage[capitalize,noabbrev]{cleveref}
\usepackage[textsize=tiny]{todonotes}
\usepackage{enumitem}
\usepackage{booktabs}
\usepackage{multirow}

\title{Chimera: State Space Models Beyond Sequences}

\author{\name Aakash Lahoti\thanks{Authors contributed equally to this work.} \email alahoti@andrew.cmu.edu \\
      \addr Carnegie Mellon University
      \AND
      \name Tanya Marwah\footnotemark[1] \email tmarwah@andrew.cmu.edu \\
      \addr Carnegie Mellon University
      \AND
      \name Ratish Puduppully \email ratishpuduppully@gmail.com \\
      \addr  IT University of Copenhagen
      \AND
      \name  Albert Gu \email agu@cs.cmu.edu\\
      \addr Carnegie Mellon University, \\
      Cartesia AI}

\newcommand{\blue}[1]{\textcolor{black}{#1}}

\def\month{04}  % Insert correct month for camera-ready version
\def\year{2025} % Insert correct year for camera-ready version
\def\openreview{\url{https://openreview.net/forum?id=yv0TUssepk}} % Insert correct link to OpenReview for camera-ready version

\newif\ifshowagcomments
\showagcommentstrue  % Use \showagcommentsfalse to hide AG comments

\newif\ifshowagrcomments
\showagrcommentsfalse  % Use \showagrcommentsfalse to hide AGR comments

\newcommand{\AGR}[1]{\ifshowagrcomments \textcolor{blue}{[AGR: #1]} \fi}

\begin{document}

\maketitle

\begin{abstract}
Transformer-based deep learning methods have emerged as the standard approach to model diverse data such as sequences, images, and graphs.
These methods rely on self-attention, 
which treats data as an unordered set of elements.
This ignores the neighborhood structure or \emph{graph topology} of the data and requires the use of inductive biases, such as position embeddings in sequences and images, and random walks in graphs, to incorporate topology. 
However, developing bespoke inductive biases for each task requires significant effort and can also introduce side-effects hindering generalization.
In this work, we introduce \emph{\frameworkname{}, a unified model that directly incorporates the data topology in a principled way}, obviating the need for domain-specific biases.
Central to \frameworkname{} is the observation that state-space models---which naturally do not require position embeddings---can be generalized to capture any general graph topology.
% \red{\frameworkname{} generalizes State-Space Models---which naturally do not require position embeddings---from causal sequences to any graph topology.}
%
Our experiments demonstrate the versatility of our approach---\frameworkname{} achieves strong performance across the domains of language, vision, and graphs, outperforming BERT on GLUE by 0.7 points, ViT on ImageNet-1k by 2.6\%, and all the baselines on the Long Range Graph Benchmark.
Our results validate \nameframework{'s} \textit{principled methodological contributions} and affirm the long-held belief that data topology is a powerful inductive bias across modalities.~We further propose \textit{algorithmic optimizations} to improve \nameframework{'s} efficiency while maintaining performance: 1) 
For the subclass of Directed Acyclic Graphs we show that \nameframework{}
can be implemented as a linear time recurrence.
2) For general graphs, we relax the method with a 
simple mathematical approximation, achieving Transformer's quadratic complexity without relying on domain-specific biases.\looseness=-1
\begingroup
\renewcommand\thefootnote{}\footnote{\noindent\textbf{Code:} \href{https://github.com/goombalab/chimera}{github.com/goombalab/chimera}}\addtocounter{footnote}{-1}
\endgroup
\end{abstract}
\section{Introduction}

Real-world data, ranging from sequential language and audio to high-dimensional images and structured molecule data, often exhibit some notion of neighborhood, or \emph{graph topology}, among its constituent elements.
For instance, language and audio have a directed line graph topology; % (Fig~\ref{fig:directed_line_graph_schema});
images possess an undirected grid-graph topology; % (Fig~\ref{fig:grid_graph_schema}); 
Structured molecule data has predefined nodes (atoms) and edges (bonds) that constitute its topology. % (Fig~\ref{fig:general_graph_schema}).
\blue{More recently}, Transformer-based~\citep{transformer} methods, with self-attention at their core, are \blue{increasingly being} used to model such data~\citep{bert,vit,rampavsek2022recipe}.
However, self-attention is permutation invariant and treats data as an unordered set of elements, disregarding its topology. 
Consequently, significant research effort has focused on developing domain-specific inductive biases, such as position embeddings~\citep{su2023roformerenhancedtransformerrotary, bert}, and random walks~\citep{graphmamba2, graphmamba}, to incorporate data topology into the model.

However, designing these inductive biases requires navigating a large search space for each domain.
For instance, RoPE embeddings~\citep{su2023roformerenhancedtransformerrotary} work well in language~\citep{touvron2023llamaopenefficientfoundation}; 
in vision, absolute and 2D-RoPE embeddings are widely used~\citep{vit,heo2024rotary}; 
while Laplacian embeddings or random walks are used in graphs~\citep{rampavsek2022recipe}. 
Moreover, these techniques can produce undesirable side effects, such as poor out-of-domain generalization---RoPE struggles to generalize to sequences longer than the training lengths~\citep{kazemnejad2024impact}, while absolute position embeddings have inherently constrained context sizes due to their design.
Furthermore, it is unclear how effectively these techniques capture the underlying graph topology.

In this paper, we introduce \emph{\frameworkname{}}, a unified 
framework that \emph{directly} incorporates data topology---i.e., the underlying graph structure---in a principled way.
\frameworkname{} is motivated by the observation that 
% \frameworkname{} builds on recent 
State Space Models (SSMs) for causal language modeling---Mamba-2~\citep{ssd}, RetNet~\citep{retnet}, and Linear Attention (LA)~\citep{la}---naturally capture the sequence order through recurrence, without position embeddings. 
We formalize this property and generalize it beyond causal sequences to any graph topology.
This approach is in contrast with existing methods that instead apply attention or 
SSMs as a black box to ``flattened data'' 
augmented with heuristics to compensate for loss of 
topological information~\citep{vmamba,vit}.
We validate our \emph{methodological innovation} with empirical results that demonstrate strong performance across diverse domains. 
This affirms the long-held belief that topology is a powerful inductive bias, while providing a principled way to incorporate it into the modeling process.

% To derive \nameframework{},
We first formally show how SSMs capture the underlying directed line graph topology of language data through recurrence (Sec \ref{subsec: mamba on a directed line graph}).
For this,
we leverage the Structured Masked Attention (SMA) representation~\citep{ssd} of SSMs: methods such as Mamba-2, RetNet, and Linear Attention are equivalent to the matrix $\mathbf{M} = \mathbf{L} \odot (\mathbf{Q}\mathbf{K}^T)$ multiplied with the input, 
where $\mathbf{Q}, \mathbf{K}$ are the query and key matrices, respectively, and $\mathbf{L}$ 
is the (data-dependent) mask matrix analogous to the causal mask in attention.
We show that the mask matrix fully encodes the topology of the underlying graph structure by acting as the \emph{resolvent} of the adjacency matrix, 
$\mathbf{A}$, of a directed line graph, i.e. $\mathbf{L} = (\mathbf{I} - \mathbf{A})^{-1} = \sum \mathbf{A}^i$, where $\mathbf{I}$ is the identity matrix. 
\AGR{be careful here, a general reader may not consider LA as obviously an SSM. think of your typical target audience and put yourself in their shoes throughout this whole paragraph; AL: We can remove LA if it seems like a distraction and replace it with some other SSM}
This result allows us to generalize to any graph topology. 
Specifically, for an ``appropriately parameterized'' adjacency matrix, $\mathbf{A}$, of the graph topology, we compute the matrix multiplication of $\mathbf{M} = \mathbf{L} \odot (\mathbf{Q}\mathbf{K}^T)$, where $\mathbf{L} = (\mathbf{I} - \mathbf{A})^{-1}$, with the input.
Intuitively, \(\mathbf{A}_{ij}\) captures the ``influence'' between neighbors $i$ and $j$, while the resolvent aggregates this influence over all paths, thus capturing the underlying topology.
In Section \ref{sec:general_graphs}, we present the detailed parameterization scheme used in Chimera which is important for both empirical performance and numerical stability.

% The primary computational 
The main
bottleneck lies in the computation of the 
mask matrix, 
% inverse operation
whose naive implementation incurs a cubic cost. 
We propose two algorithmic optimizations to mitigate this cost while maintaining performance: 

\AGR{this should probably be an enumerate with margin-reducing options, or just in english "First" "Second"}
\begin{enumerate}[topsep=0pt, parsep=0pt, itemsep=2pt, partopsep=0pt, left=2pt]
    \item We specialize the method for the subclass of directed acyclic graphs (DAGs). 
    This is motivated by the fact that many graph topologies can be canonically decomposed into multiple DAGs. 
    For example, an undirected line graph can be decomposed into two directed line graphs (Fig \ref{fig:directed_line_graph_decomposition}), 
    while a grid graph can be divided 
    into four directed grid graphs (Fig \ref{fig:2D_grid_decomposition}).
    We prove that for this subclass, the resolvent can be computed by running a recurrence that is \blue{linear in the number of nodes and edges.}
    We further propose a squaring technique to compute the resolvent efficiently on modern hardware accelerators by leveraging matrix multiplications, at a quadratic cost \blue{in the number of vertices. This cost is optimal in the worst case, it can be improved when the underlying graph is``structured''---line graphs can computed in linear time via existing Mamba-2 kernels} 

    \item We relax the exact computation of the resolvent for general graphs with a finite sum approximation of the Neumann series, i.e. $(\mathbf{I} - \mathbf{A})^{-1} = \sum_{i=0}^d \mathbf{A}^i$, where $d$ is the diameter of the graph. 
    We can efficiently compute this approximation with a squaring technique, 
    capturing the global topological structure.
    We further show that the finite sum approximation performs 
    as well as the method with the sum of infinte terms. 
\end{enumerate}

\AGR{may warrant a comment on the usefulness of this: covers some of the most important cases of sequences and images}

Overall, we make the following contributions:\looseness=-1
\begin{itemize}[topsep=0pt, parsep=0pt, itemsep=2pt, partopsep=0pt, left=2pt]
    \item 
    % \red{I think we shouldn't lead with that we are first. we should end with, to the best of our knowledge, chimera is the 
    % first model that generalizes ssms and models different topologies...}
    We propose \nameframework{}, 
    a unified model that directly incorporates graph topology in a principled way by generalizing SSMs. This is in contrast with existing approaches that apply attention or SSMs as a black box on ``flattened data'' with additional heuristics.
    % \item We introduce algorithmic optimizations to improve the method's efficiency by specializing it to DAGs and approximating the resolvent using a finite sum, while preserving performance. We show that for canonical modalities such as images and language, where data can be decomposed into DAGs, this finite-sum approximation becomes exact.
    \item We introduce algorithmic optimizations to improve the method's efficiency by specializing it to DAGs and approximating the resolvent using a finite sum, while preserving performance. We show that for canonical modalities such as images and language, where data can be decomposed into DAGs, this finite-sum approximation becomes exact.
    \item We demonstrate that \frameworkname{} consistently achieves strong performance across diverse domains including language, images, and graphs---outperforming BERT~\citep{bert} with a GLUE score~\citep{glue} of 0.7, surpasses ViT~\citep{vit} on ImageNet-1k~\citep{imagenet} classification by 2.6\%. Furthermore, our method
    outperforms strong baselines on the Long Range Graph Benchmark (LRGB)~\citep{dwivedi2022long}, 
    where we show that our model is capable of modeling both long and short range 
    interactions nodes, while respecting the graph structure.
\end{itemize}

% \begin{figure*}[!t]
% \vspace{-0.7cm} 
%   \centering
%   \begin{subfigure}[b]{0.2\textwidth} % Adjusted width
%     \centering
%     \includegraphics[scale=0.55,keepaspectratio]{Images/sentence_line.png}
%     \caption{Language (Line Graph)}
%     \label{fig:directed_line_graph_schema}
%   \end{subfigure}
%   \hfill
%   \begin{subfigure}[b]{0.2\textwidth} % Adjusted width
%     \centering
%     \includegraphics[scale=0.22,keepaspectratio]{Images/cat_2d.png}
%     \caption{Images (Grid Graph)}
%     \label{fig:grid_graph_schema}
%   \end{subfigure}
%   \hfill
%   \begin{subfigure}[b]{0.2\textwidth} % Adjusted width
%     \centering
%     \includegraphics[scale=0.4,keepaspectratio]{Images/molecule.png}
%     \caption{Molecules (General Graph)}
%     \label{fig:general_graph_schema}
%   \end{subfigure}
%   \caption{Real-world data exhibits inherent topology: (a) language follows a directed line graph, (b) images a grid graph, and (c) structured data like molecules have explicit graph topology.}
%   \vspace{-11pt}
% \end{figure*}

\section{Preliminaries}
\label{sec:prelims}
We introduce State Space Models (SSMs), which are recurrent models designed to process sequential data, such as language and audio.
We formulate SSMs in their recurrent form and then introduce the Structured Masked Attention (SMA)~\citep{ssd} representation that unrolls and vectorizes this recurrence as a matrix $\mathbf{M}$ acting on the input $\mathbf{X}$.  
This SMA representation would allow us to show that SSMs inherently operate on a directed line graph topology.

\subsection{Overview of State Space Models}
\label{sec:prelims mamba2}
SSMs, such as Mamba-2~\citep{ssd},  Linear Attention (LA)~\citep{la}, RetNet~\citep{retnet}, are recurrent sequence-to-sequence models that feature a linear hidden-state transition function.
%
% This linearity enables a hardware-efficient, vectorized implementation of SSMs, allowing them to scale effectively.
% DO NOT MENTION THIS, as it goes against our 
%
This function is typically data-dependent which is known to improve model performance \citep{hwang2024hydra}. 

Formally, let $\bX \in \mathbb{R}^{T \times D}$ denote the input sequence of $T$ tokens, where each token has $D$ channels.
Let the size of the hidden state be $d$. 
Let $\bY \in \mathbb{R}^{T \times D}$ be the output of the sequence-to-sequence model.
Then, SSMs first compute the following matrices:
\begin{align}
    \bB = f_B(\bX), \hspace{1pt}
    \bC = f_C(\bX), \hspace{1pt}
    \bV = f_V(\bX) \in \mathbb{R}^{T \times d},
\end{align}
where $f_B$, $f_C$, $f_V$ are model specific data dependent functions. 
For instance, in Mamba-2 each of these functions is a composition of a 
linear projection of $\bX$ along the channel dimension, followed by a short convolution layer along the sequence dimension and a Swish activation function~\citep{ramachandran2017searching}.
In \citet{ssd}, it was shown that
we can view the $\bB$, $\bC$, $\bV$ matrices as analogs of the key, query, value matrices in self-attention, respectively.

Let $\vv^i =\bV[:,i] \in \mathbb{R}^{T}$ denote the input corresponding to channel $i$.
Let $\bB_t = \bB[t,:]$, $\bC_t = \bC[t,:]$ for any time $t$. 
Let $y_t^i = \bY[t,i]$ and $v^i_t = \vv^i[t]$. 
Then, the model computes the following recurrence,
starting with the hidden state vector $\vh^i_{-1} = \mathbf{0} \in \mathbb{R}^{d}$:
\begin{align}
    \vh^i_{t} &= a_t\vh^i_{t-1} + b_t \bB_t v^i_t, \\
    y^i_{t} &= \bC_t^{T} \vh^i_{t},
    \label{eq: mamb2 recurrence}
\end{align}
where $a_t,b_t$ are model-specific parameters that characterize the SSM. 
LA sets $a_t=b_t=1$, RetNet sets $a_t=\gamma$, $b_t=1$ for a learnable parameter $\gamma$.
In contrast, Mamba-2 sets $a_t,b_t$ in a data-dependent manner that implicitly encodes a gated memory mechanism known as \emph{selectivity} or the \emph{selection mechanism}.  
This allows the model to select and propagate important tokens across long sequences. Specifically,
\begin{align}
    \Delta = f_\Delta(\bX) \in \mathbb{R}^{T}; \hspace{1pt}
    a_t = \exp(-\Delta_t),
    b_t = \Delta_t \in \mathbb{R},
\end{align}
where $\Delta$ is the selectivity matrix, \blue{$\Delta_t$ is the $t^{\text{th}}$
element of the vector,}
and $f_\Delta$ like $f_B$, $f_C$, $f_V$ is a data-dependent function. 
Selectivity works by assigning larger values $\Delta_t$ to important tokens, amplifying their contribution to the previous hidden state, while assigning smaller values $\Delta_t$ to unimportant tokens, which preserve the past hidden state with minimal influence from these tokens. 

\subsection{The Structured Masked Attention Representation}
\label{sec: prelim sma}

\citet{ssd} introduced the Structured Masked Attention (SMA) representation for SSMs, which vectorizes the time-stepped recurrence (Eq.~\ref{eq: mamb2 recurrence}) as a matrix multiplication, $\bY = \bM \bV$.
\footnote{%
Not all SSMs admit an SMA representation. We focus on those that do, such as LA, RetNet, and Mamba-2. In this work, we use the term ``SSMs'' specifically to refer to this restricted class.%
}
Here, $\bM$ depends on the data-driven matrices $\bB, \bC,$ and $\Delta$, and can be expressed as $\bM = \bL \circ \bigl(\bC\bB^T\bigr)$, where $\bL$ is a mask derived from $\Delta$. 
One can obtain this formulation by unrolling the recurrence across all time steps.

Formally, define $\bar{\bB}_t = b_t \bB_t$, and recall from Section \ref{sec:prelims mamba2} that $b_t = \Delta_t$, $a_t = \exp(-\Delta_t)$ for Mamba-2; $b_t = 1$, $a_t = \gamma$ for RetNet; and $b_t = 1$, $a_t = 1$ for Linear Attention.
Then the output $\bY$ of the recurrence (Eq. \ref{eq: mamb2 recurrence}) can be vectorized as,
\AGR{also, the $\bM$ notation is introduced but never used again. either drop it or in equation (5) say $y = Mv = (L \circ CB^T) V$.}
\begin{align}
    \bY = \bM\bV = (\bL \odot \bC\bar{\bB}^T)\bV,
    \label{eq: sma}
\end{align}
where the mask matrix $\bL_{ij} = \mathbf{1}[i \ge j] \: \Pi_{j < k \le i} a_k$,
% \begin{align}
%     \bL =  
%     \begin{bmatrix}
%         1 & 0 & \cdots & 0 \\
%         a_1 & 1 & \cdots & 0 \\
%         a_1 a_2 & a_2 & \cdots & 0 \\
%         \vdots & \vdots & \ddots & \vdots \\
%         a_1 a_2 \cdots a_{T-1} & a_2 a_3 \cdots a_{T-1} & \cdots & 1
%     \end{bmatrix}.
%     \label{mamba_L_matrix}
% \end{align}
The SMA representation of the recurrence (Eq. \ref{eq: mamb2 recurrence}) is useful because it neatly isolates the effect of the underlying topology within the recurrence computation into the mask matrix $\bL$ (Sec. \ref{sec: sec3})
This property will allow us to generalize SSMs to arbitrary topologies by appropriately formulating the structured mask $\bL$.

\section{\frameworkname{}: Incorporating Graph Topology}

\label{subsec: mamba on a directed line graph}
\begin{figure*}[t!]
\includegraphics[width=1.0\textwidth]{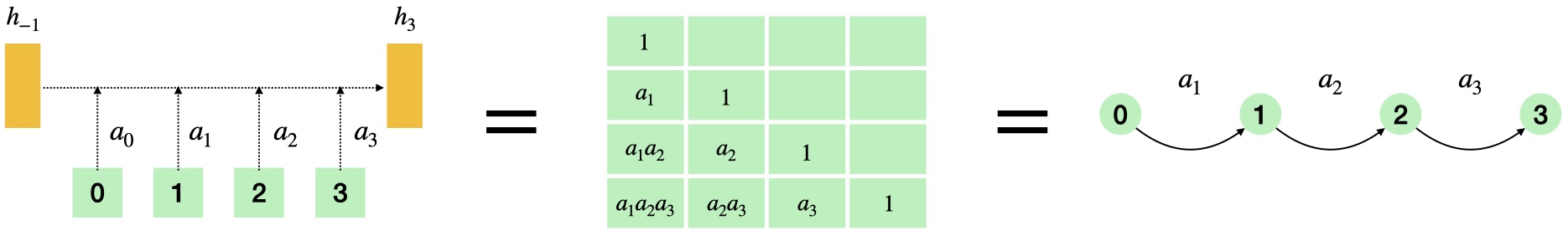}
\caption{\emph{SSMs inherently operate on a directed line graph topology}: SSMs modeling a sequence of tokens in the recurrent representation (left), the structured mask matrix from the SMA representation of SSMs (center), The underlying directed line graph topology (right).}
\label{mamba_is_line_graph}
\end{figure*}

\label{sec: sec3}
In this section, we introduce \frameworkname{}, a unified model that directly incorporates the underlying graph topology of a domain by mathematically generalizing SSMs.
This contrasts to existing methods, such as \citet{graphmamba2, bert, swin}, that use attention or SSMs as a black-box applied to `flattened data' and rely on inductive biases to incorporate structural information.  

Our motivation stems from the fact that SSMs on causal language modeling task do not require position embeddings and naturally capture the sequence order with their recurrence. 
We seek to formalize this result which then allows us to generalize it to arbitrary graphs.
To this end, we begin by defining the resolvent of a linear operator and interpret its action when this operator is the adjacency matrix.

\subsection{Resolvent of an Adjacency Matrix Accumulates Influence Along All Paths}
\label{sec: resolvent}
A graph consists of a set of nodes \(\gV\) that represent data elements, and edges \(\gE\) that encode the underlying topological structure.
We conceptualize the associated adjacency matrix \(\bA \in \mathbb{R}^{|\gV| \times |\gV|}\) as \emph{capturing the influence between neighboring nodes}. Specifically, \(\bA_{ij}\) is the influence that node \(j\) has on node \(i\), for each edge \((i,j)\).
The desideratum is to extend the notion of influence to all node pairs by incorporating the graph's structure, accounting for all possible paths between them.
To model this cumulative influence, we introduce the concept of the resolvent of a linear operator

\begin{definition}[Resolvent of a Linear Operator~\citep{reed1980methods}]
Let $\bA \in \mathbb{R}^{T \times T}$ be a linear operator, $\bI$ the identity operator, and $\lambda$ a complex number. Then, the resolvent operator is defined as:
\begin{align}
    R(\lambda, \bA) = (\lambda \bI - \bA)^{-1},
\end{align}
which exists for all complex numbers $\lambda$ that are not in the spectrum of $\bA$, i.e., $\lambda \notin \sigma(\bA)$. 
In this work, we set \(\lambda = 1\) to remain in the field of real numbers, and this is done without loss of generality, as any choice of $\lambda$ is equivalent upto scaling of the model.
\end{definition}

We now demonstrate how the resolvent operator captures the influence between any two nodes in the graph. 
Observe that the resolvent operation can be expanded using the Liouville-Neumann series if the operator norm of the adjacency matrix is strictly less than 1, i.e.  \(\|\bA\| < 1\),
\begin{align}
    R(1, \bA) = (\bI - \bA)^{-1} = \sum_{k=0}^{\infty} \bA^k.
    \label{eq:neumann}
\end{align}
Intuitively any term, \(\bA_{ij}^k\), in this expansion represents the influence between nodes \(i\) and \(j\) accumulated across all paths of length exactly \(k\) connecting them. We formalize this in the following Proposition \ref{prop: A^k}.

\begin{proposition}[\(\bA^k\) accumulate influence through paths of length $k$]
\label{prop: A^k}
Given the weighted adjacency matrix $\bA \in \mathbb{R}^{T \times T}$ of a graph $\gG = (\gV, \gE)$ with $|\gV| = T$, \blue{where $A_{ij}$ is the weight of the $(i, j)$ edge of the graph.}
Then $(i, j)^{\text{th}}$ entry of \(\bA^k\) is given as,
\[
(\bA^k)_{ij} = \sum_{p_1, p_2, \dots, p_{k-1}} \bA_{ip_1} \bA_{p_1p_2} \cdots \bA_{p_{k-1}j},
\]
where $(p_1, \dots, p_{k-1})$ is an ordered sequence of vertices forming a path of length \(k\) from node \(i\) to \(j\).
\end{proposition}

Therefore, the series \((\bI - \bA)^{-1}_{ij}\) (Eq. \(\ref{eq:neumann}\)) sums up the influence of node \(i\) on node \(j\) over all possible lengths. 
% Additionally, we also note that Eq. $\ref{eq:neumann}$ provides a sufficient condition for the existence of the resolvent: the series converges when the operator norm of \(\bA\) is less than one.

\subsection{SSMs operate on a Directed Line Graph}

We now show that SSMs naturally operate on a directed line graph. 
Specifically, let \(\gV\) be the set of tokens, and \(\gE\) be the edges connecting token \(t\) to the next token \(t+1\). 
Let the weighted adjacency matrix be \(\bA_{s, t} = \mathbf{1}_{[t=s+1]} a_t\), where \(a_t\) is the SSM-specific parameter defined in Section \ref{sec: prelim sma}. 
% For convenience, we recall that for Mamba-2, \(a_t = \exp(-\Delta_t)\), where \(\Delta_t\) is the selectivity parameter; for RetNet, \(a_t = \gamma\), which is a constant decay factor; and for Linear Attention, \(a_t = 1\).

Recall from Section \ref{sec: prelim sma} that SSMs are equivalent to the matrix action of \(\mathbf{M} = \mathbf{L} \odot (\mathbf{C}\mathbf{B}^T)\) on the input. 
We make the key observation that \emph{\(\bL\) is precisely the resolvent of \(\bA\)}, that is $\bL = (\bI - \bA)^{-1}$.
This mathematically ties SSMs' recurrence to the directed line graph topology, with the mask encoding the topology (Fig \ref{mamba_is_line_graph}).

\begin{proposition}
\label{prop:mamba_is_dlg}
    Under the notation established in Section \ref{sec:prelims}, consider a weighted directed graph $\gG$ with nodes $\gV = \{0, \cdots, T-1\}$, edges $\gE = \{ (i-1, i) | i \in \gV, i > 0\}$, and the edge weights 
    $\gW = \{ w_{\blue{(i-1,i)}} = a_i | i \in \gV, i > 0\}$.
    Let $\bA$ be the weighted adjacency matrix of incoming edges,
    \begin{align}
    \bA =  \begin{bmatrix}
        0 & 0 & 0 & \cdots & 0 \\
        a_1 & 0 & 0 & \cdots & 0 \\
        0 & a_2 & 0 & \cdots & 0 \\
        \vdots & \vdots & \vdots & \ddots & \vdots \\
        0 \cdots 0 & 0 \cdots 0 & 0 & a_{T-1} & 0
    \end{bmatrix},
\end{align}
    then $\bL = \sum_{i=0}^\infty \bA^i = (\bI-\bA)^{-1}$, 
    and consequently,
    $ \vy = ( (\bI-\bA)^{-1} \odot \bC\bar{\bB}^T)\bV$.
\end{proposition}

We interpret this result intuitively: In a directed line graph, there is exactly one path between the tokens \(i\),\(j\) with \(i < j\), and the corresponding mask matrix entry \(\bL_{ij} = \prod_{i \geq k > j} a_k\), reflects the cumulative influence of the intervening tokens along this path. Furthermore, \(\bL_{ij} = 0\) for \(i < j\) restricts influence in the forward direction, ensuring causality. This shows that SSMs inherently operate on a directed line graph with the $\bL$ matrix encoding the topology.

\subsection{Generalizing SSMs to Arbitrary Graph Topologies}
\label{sec:general_graphs}
Building on Proposition \ref{prop:mamba_is_dlg}, we can generalize SSMs from causal sequences to arbitrary graph topologies.
Specifically, we compute the resolvent of the adjacency matrix, \(\bA\), with output \(((\bI - \bA)^{-1} \odot (\bC\bar{\bB}^T)) \bV\).

We focus on the parameterization of \(\bA\) with the following key points: 1. It ensures the numerical stability of the method by addressing cases of non-invertibility or poor conditioning of resolvent; 
2. It generalizes Mamba-2's selectivity that allows for modeling long-range dependencies.\footnote{Our approach applies to any SSM with an SMA representation and in this work, we specifically use Mamba-2.}

Formally, consider a graph \(\gG = (\gV, \gE)\) with \(|\gV| = T\) nodes, where each node has \(D\) channels. Let \(d\) denote the generalized hidden state size. 
For each node, we compute,
\begin{align}
    &\bB = f_B(\bX),
    \bC = f_C(\bX),
    \bV = f_V(\bX) \in \mathbb{R}^{T \times d}, \\
    &\Delta = f_\Delta(\bX) \in \mathbb{R}^{T}, \label{eq:f_delta}
\end{align}

where the functions \(f_B\), \(f_C\), $f_V(\bX)$, \(f_\Delta\) are linear projections applied to the input, followed by a local graph convolution over neighboring nodes and a Swish activation as chosen in Mamba-2. 
Furthermore, if the data set features edge embeddings $\mathbf{E} \in \mathbb{R}^{|\mathcal{E}| \times D}$, we define 
% $\Delta^E = f_{\Delta'}(\mathbf{Z}) \in \mathbb{R}^{|\mathcal{E}|}$, 
\blue{$\Delta' = f_{\Delta'}(\mathbf{Z}) \in \mathbb{R}^{|\mathcal{E}|}$, where $\mathbf{Z} \in \mathbb{R}^{D}$ is an edge embedding},
as the selectivity matrix corresponding to the edges.
\blue{Here $f_{\Delta'}$ is computed similarly to $f_\Delta$ as in \eqref{eq:f_delta}.}

We parameterize the \(\bA\) matrix for each edge \((i,j) \in \gE\) as,
\begin{align}
    \bA_{ij} = \exp\left(
        -\frac{\Delta_i + \Delta_j + \Delta'_{(i,j)}}{3}
    \right),
\end{align}
% \textcolor{red}{this is incorrect, we have two 
% deltas, for forward and backward.}
to incorporate context from both ends of the edge $(i,j)$ as well as the edge embeddings.
\blue{Here $\Delta_i$, $\Delta_j$ and $\Delta'_{(i, j)}$ are 
are the learned selectivity parameters for the nodes $i$, $j$ and edge $(i, j)$ respectively.}
To add directionality to $A_{ij}$ and to further increase the representational power of our model, 
we can also maintain two (different) $\Delta$'s such that 
$
\bA_{ij} = \exp\left(
    -\frac{\Delta^{(1)}_i + \Delta^{(2)}_j + \Delta'_{(i,j)}}{3}
\right)
$.

Note that the matrix \(\bI - \bA\) may be non-invertible or poorly conditioned, which would inhibit inverse computation and stable training of the model.
We mitigate this issue with a data-dependent normalization parameter \(\Psi = f_\Psi(\bX) \in \mathbb{R}^{T}\), computed similarly to \(\Delta\), and perform a row-wise normalization of the adjacency matrix using \(\Psi\). 
Specifically, for each row \(i \in [T]\), we apply:
\[
\bA[i,:] = \frac{\gamma \bA[i,:]}{\mathbf{1}^T\bA[i,:] + \exp(-\Psi_i)},
\]
where \(\gamma\) is a scaling hyperparameter. 
The following proposition shows that this normalization guarantees the convergence of the Neumann series for the adjacency matrix $\bA$. 
\begin{proposition}
\label{prop:row_wise_normalization}
    Under Gaussian initialization, the row-wise normalization strategy ensures that \(\|\bA\| < 1\) and \(\|(\bI - \bA)^{-1}\|\) 
    is bounded with probability
    \blue{$> 1-\Phi(\tfrac{-1}{2\gamma})$}.
\end{proposition}
The proof for this proposition in Appendix~\ref{proof:row_wise_normalization}.
\AGR{This appendix link seems to be broken?}
Finally, we compute the resolvent matrix $\bL = (\bI - \bA)^{-1}$ and the output $\vy$ as $(\bL \odot \bC \bar{\bB}^T) \bV$.

\section{\frameworkname{} With Improved Efficiency}
\label{sec: efficient_mamba}

While \frameworkname{} supports arbitrary graph topologies, computing the resolvent incurs a cubic cost in the number of nodes, which can be prohibitively expensive for large graphs. 

In this section, we propose two algorithmic optimizations to mitigate this cost:
First, we specialize \frameworkname{} to a tractable yet expressive subclass of Directed Acyclic Graphs (DAGs) for which the resolvent can be computed in linear time by running a recurrence on the topologically sorted graph.
Second, for general graphs, we relax the resolvent computation using a finite approximation, achieving quadratic complexity of Transformers without domain-specific heuristics. 

\begin{figure}[b]
    \centering
    \includegraphics[width=0.7\linewidth]{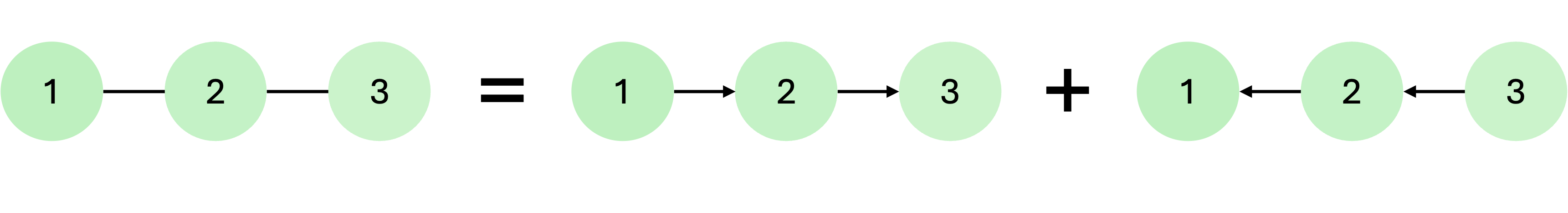}
    \vspace{-0.5cm}
    \caption{Canonical DAG decomposition of undirected line graph topology (left) into two directed line graph topologies (right).}
    \label{fig:directed_line_graph_decomposition}
\end{figure}

\begin{figure}[b]
    \centering
    \vspace*{-0.5em}
    \includegraphics[width=0.99\linewidth]{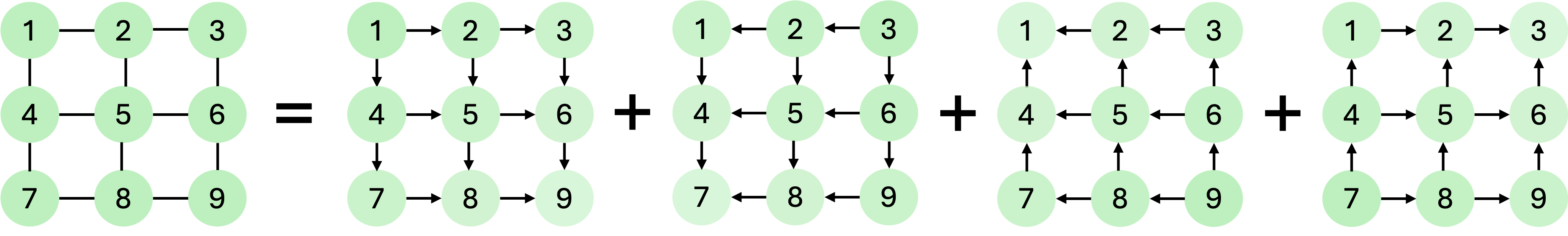}
    \vspace*{-0.5em}
    \caption{
        Grid graph (left). The canonical 2D-DAG decomposition of the grid graph (right). 
        These graphs are sufficient to capture the influence between all pairs of nodes in the undirected grid graph.
        % \AG{Should explain why we need 4 graphs, so all pairs of dependencies are covered}
    }
    \vspace*{-0.8em}
    \label{fig:2D_grid_decomposition}
\end{figure}
\subsection{\frameworkname{} on DAGs}
In this section, we introduce a tailored normalization scheme as well as a linear-time recurrent algorithm for \nameframework{} on DAGs.
We further propose a modern accelerator-friendly technique to compute this resolvent efficiently by leveraging matrix multiplications, although at the cost of quadratic FLOPs.

Our choice of the DAG subclass is motivated by its expressivity.
Topologies such as undirected line and grid graphs can be canonically decomposed into DAGs:
line graph divides into two directed line graphs (Fig \ref{fig:directed_line_graph_decomposition}) and grid graph divides into four directed grid graphs (Fig \ref{fig:2D_grid_decomposition}). 
This allows for efficient \frameworkname{} that preserves the grid topology of an image.

\subsubsection{\frameworkname{} on DAGs: The method}
\label{sec:mamba-dag}
Formally, consider a DAG, \(\gG = (\gV, \gE)\), with \(|\gV| = T\) nodes, each with \(D\) channels and a hidden state size of \(d\). 
For any node \(i\), let \(p(i)\) be the set of its parents. 
Let $\bB, \bC, \bV, \Delta$ be the input projections as defined in Section \ref{sec: sec3}.
\blue{We define the adjacency matrix \(\bA\) as \(\bA_{ij} = \exp(-\Delta_{ij})\) for each \((i,j) \in \gE\), and set $\bar{\bB}_i = (\sum_{(i,j)\in\gE}\Delta_{ij})\bB_i$ for each node $i$. 
In this work, we define $\Delta_{ij} = (\Delta_{i}+\Delta_{j})/2$, but more generally one can take
$\Delta_{ij} \;=\;f(\Delta_i,\,\Delta_j)$
for any suitable (e.g.\ symmetric) function $f$ of the nodewise parameters.} Then the output $\vy = (\bL \odot (\bC \bar{\bB}^T)) \bV$.

We first show that the resolvent \((\bI - \bA)^{-1}\) exists.

\begin{figure}
    \centering
    \includegraphics[width=0.5\linewidth]{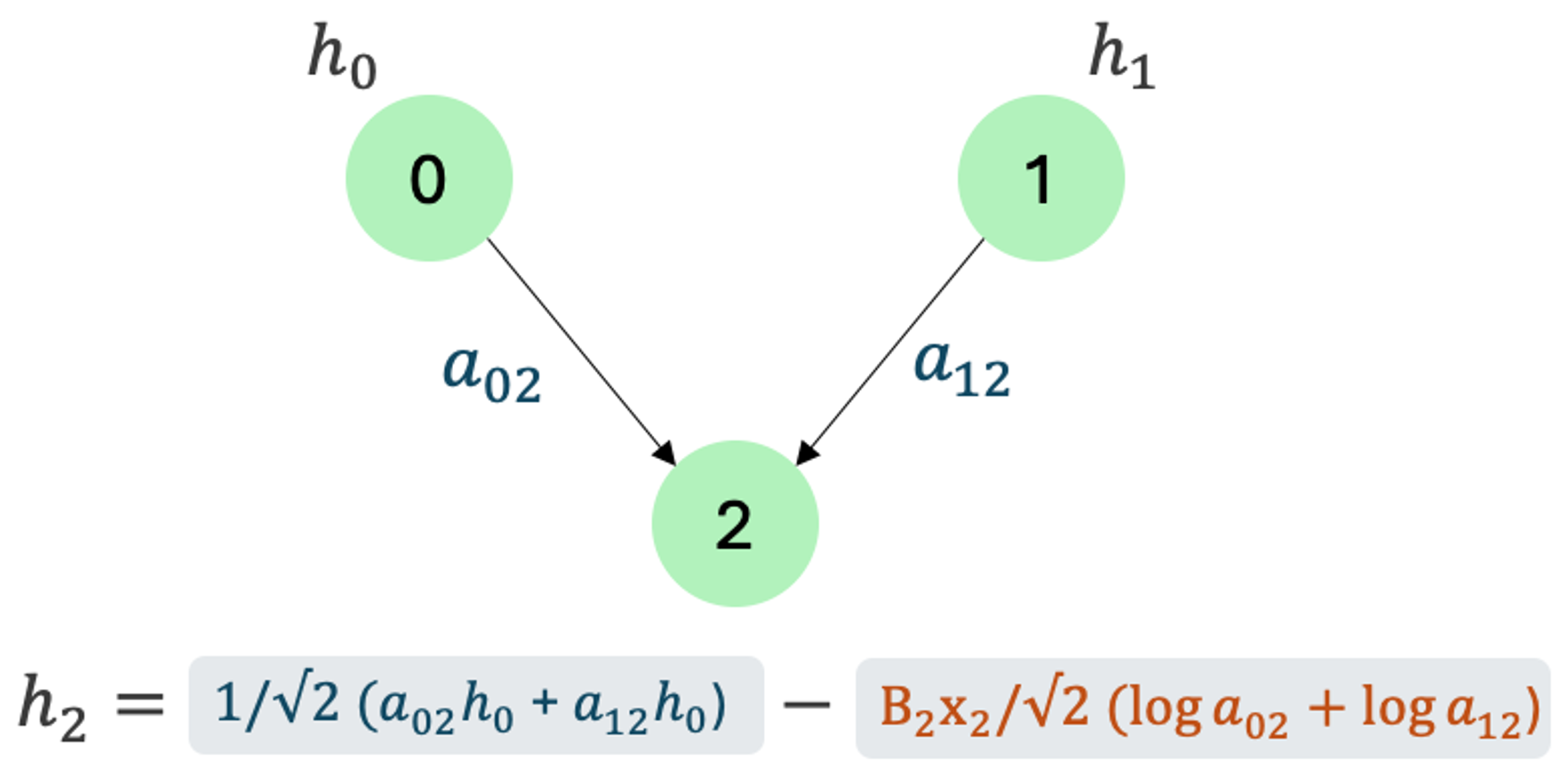}
    \caption{\nameframework{} on DAGs: A visualization of the normalized recurrence. The absence of cycles in DAGs enables a recurrent view of the method which allows for a fast linear-time computation.}
    \label{fig:dag-recurrence}
\end{figure}
\begin{proposition}
\label{prop:nilpotent}
For a DAG, \(\bA\) is nilpotent, that is 
\blue{\( \bA^K = \mathbf{0} \)}.
Therefore, the inverse \((\bI - \bA)^{-1}\) exists and is given by the finite sum:
\begin{align}
    \bL = (\bI - \bA)^{-1} = \blue{\sum_{k=0}^{K-1} \bA^k}.
\end{align}
\end{proposition}
While the resolvent always exists, we note that its entries can become exceedingly large which can cause numerical instabilities. 
To show this, we first represent this method in its recurrent view in Prop. \ref{prop:graph_recurrence}
\blue{(and is visualized in Figure~\ref{fig:dag-recurrence})}.
\begin{proposition}
\label{prop:graph_recurrence}
Our method computes the following recurrence on each 
% channel $\vv$ of $\bV$:
\blue{channel $v_i$ of $\bV$:}
    \begin{align}
        \vh_i = 
        \sum_{j \in p(i)} 
        \bA_{ij} \vh_j -  \bar{\bB}_i v_i,
        \hspace{0.5cm}
        y_i = \bC_i^T\vh_i,
    \end{align}
where $\vh_l = \mathbf{0}$ for all leaf nodes $l$.
\end{proposition}

Recall from Section \ref{sec: resolvent} that each \(\bL_{ij}\) represents the cumulative sum of all paths from node \(j\) to \(i\), and in the worst case, the number of such paths and its resolvent entry grows exponentially with distance.
To address this, we introduce a normalization scheme built directly into the recurrence:
\begin{proposition}
\label{prop:dag_normalizer}
The normalized method is:
    \begin{align}
        \vh_i& = 
        \frac{1}{\sqrt{|p(i)|}}
        \sum_{j \in p(i)} 
        \left(
            \bA_{ij} \vh_j - \ln(\bA_{ij}) \bB_i v_i 
        \right),
        \\
        y_i &= \bC_i^T\vh_i.
    \end{align}
    This normalization ensures that $\text{Var}(\bC_i^T \vh_i) \le 1$ under the assumption that the vectors \(\{\bB_i v_i, \bC_i \}_i\) are i.i.d. Gaussians, that is \(\bB_i v_i, \bC_i \sim \gN(\mathbf{0},\bI_d)\).
\end{proposition}
% \begin{figure}[b]
%     \centering
%     \includegraphics[width=0.7\linewidth]{Images/undirected_to_directed.png}
%     \vspace{-0.5cm}
%     \caption{Canonical DAG decomposition of undirected line graph topology (left) into two directed line graph topologies (right).}
%     \label{fig:directed_line_graph_decomposition}
% \end{figure}
\blue{
The assumption 
that for all $i$ we have $\bB_i, v_i, \bC_i \sim \gN(\mathbf{0}, \bI_d)$
is justified given that weights 
are usually initialized with zero-mean 
and scaled-identity covariance (Xavier initialization~\cite{glorot2010understanding}), 
in addition to the fact that such distributions are approximately preserved throughout training via normalization techniques.
}
The proof follows by induction on the time step \(t\), ensuring that the output variance is bounded by 1, \(\text{Var}(\bC_i^T \vh_i) \leq 1\). The detailed proof in Appendix~\ref{proof:dag_normalizer}. 
To incorporate this normalization in the SMA representation, we define,
\begin{align}
    \bar{\bA} = \tfrac{1}{\sqrt{|p(i)|}} \bA,  \hspace{10pt} 
    \bar{\bB} = \tfrac{\ln(\bA_{ij})}{\sqrt{|p(i)|}} \bB,  \hspace{10pt} 
    \bL = (\bI - \bar{\bA})^{-1},
\end{align}
and compute the output $\vy = (\bL \odot (\bC \bar{\bB}^T)) \bV$.

\subsubsection{\frameworkname{} is efficient on DAGs}
\label{sec:chimera_dag_efficiency}
Finally, we highlight that DAGs are a particularly important case of \frameworkname{} because of additional efficiency benefits, both through recurrent and vectorized implementations.

% \red{Recurrent View: Linear Time Complexity. Should we write the explicit linear dynamics that the recurrent view taht gives the linear complexity??}
\paragraph{Linear-Time Complexity in the Recurrent View}
The intuition for linear complexity is that the resolvent operation for DAGs is \emph{finite} because of the lack of cycles.
From the adjacency matrix perspective, $\bA$ is nilpotent, i.e. $\bA^k = 0$, where $k$ is the diameter of the graph (Prop ~\ref{prop:nilpotent}).
Thus, when running \frameworkname{} as a recurrence on the DAG, the resolvent operation converges after one pass over the graph in the topologically-sorted order, which takes linear time.

% \red{However, the recurrent view is not amenable to matrix multiplication? already done in the next section}

\begin{proposition}
\label{prop:linear}
The \frameworkname{} structured mask matrix $L$ can be computed in $O(|\gV+|\gE|)$ complexity where $|\gV|, |\gE|$ is the number of vertices and edges of the graph, respectively.
\end{proposition}
The proof is provided in Appendix~\ref{proof:linear}.
We note that the linear-time complexity of Mamba-2 can be seen as a special case of Proposition \ref{prop:linear} specialized to the directed line graph, where both $|\gV|$ and $|\gE|$ is equal to the sequence length.

\paragraph{Improving Efficiency Through Matrix Multiplications}
Finally, we note that on modern hardware accelerators such as GPUs and TPUs,
various computational algorithms can have different efficiency tradeoffs.
For example, on directed line graphs, the naive computation of SSMs and RNNs as a recurrence is not parallelizable and is inefficient in practice~\citep{mamba}.
In the case of DAGs, we present a technique to reduce both the forward and backward pass for \nameframework{} to leverage only matrix multiplications which are heavily optimized on modern accelerators.
Although this technique is highly parallelizable, it requires the materialization of the adjacency matrix which is quadratic in the number of nodes, $|\gV|$.
\blue{In the worst-case, i.e. a complete transitive DAG with $\lvert E\rvert = \tfrac{|\gV|(|\gV|-1)}{2}$), this bound is tight. 
However, for specialized structured subclasses (e.g., line graphs) where the number of edges is $O(\gV)$, this cost can be significantly reduced (e.g. via existing Mamba-2 kernels for line graphs).}
The proof for the following theorem is in Appendix \ref{proof:dagthm}.
\begin{theorem}
\label{thm:dagthm}
In case of \frameworkname{} on DAGs, the forward pass can be computed with $O(\log(\text{dia}(\gG)))$ matrix multiplications where $\text{dia}(\gG)$ is the diameter of the graph (i.e. length of the longest path), and the backward pass can be computed with $O(1)$ matrix multiplications.
\end{theorem}

% \begin{figure}[t]
%     \centering
%     \vspace*{-0.5em}
%     \includegraphics[width=0.99\linewidth]{Images/2DGridDecomp.png}
%     \vspace*{-0.5em}
%     \caption{
%         Grid graph (left). The canonical 2D-DAG decomposition of the grid graph (right). 
%         These graphs are sufficient to capture the influence between all pairs of nodes in the undirected grid graph.
%         % \AG{Should explain why we need 4 graphs, so all pairs of dependencies are covered}
%     }
%     \vspace*{-0.8em}
%     \label{fig:2D_grid_decomposition}
% \end{figure}

\begin{table*}[t]
\caption{
Comparing \nameframework{} on the undirected line graph (UG), and on DAG decomposed directed line graphs (DAG) with other state-of-the-art models including M2~\citep{m2}, MLP-Mixer~\citep{mlpmixer}, FNet~\citep{fnet}, BERT~\citep{bert} on GLUE benchmark.
Chimera outperforms all baselines including BERT with a linear time complexity. 
\blue{Here the best numbers are highlighted in \textbf{bold}
and the second best numbers for each task 
are \underline{underlined}}.}
\centering
\small
\resizebox{\linewidth}{!}
{ %< auto-adjusts font size to fill line
\begin{tabular}{lcccccccccccc} %@{}}
    \toprule
    \multirow{2}{*}{Method} & \multirow{2}{*}{\#Params} & \multicolumn{2}{c}{Pretrain} & \multicolumn{8}{c}{GLUE Tasks} & \multirow{2}[2]{*}{\begin{tabular}[c]{@{}c@{}}GLUE\\Avg\end{tabular}} \\ % \multirow{2}{*}{\shortstack{GLUE\\Avg}}
    \cmidrule(lr){3-4}\cmidrule(lr){5-12}
                                            & & $\mathcal{L}_{ce}$  & Acc (\%) & MNLI & QNLI  & QQP   & RTE   & SST2  & MRPC  & COLA  & STS \\
    \midrule
    BERT-Base        & 110M  & 1.59 & 67.3 & \underline{84.1}  & \textbf{89.8}  & \textbf{91.2}  & \underline{77.2}  & 91.2  & 87.5  & 54.6  & \textbf{88.9}  & \underline{83.2}  \\
    MLP-Mixer   & 112M  & 1.77  & 63.5  & 77.2  & 82.4  & 87.6  & 67.3  & 90.5  & 86.5  & 43.0  & 85.2  & 77.5  \\
    FNet        & 112M  & 1.94 & 61.3 & 74.9 & 82.1 & 85.7 & 63.6 & 87.6 & 86.4 & 42.7 & 83.1 & 75.8 \\
    M2          & 116M  & 1.65  & 65.9  & 80.5  & 86.0  & 87.0  & 69.3  & 92.3  & 89.2  & 56.0  & 86.9  & 80.9  \\
    \nameframework{} (UG)  & 110M & 1.49 & \underline{68.5} & 83.63 & 88.98 & 89.32 & 73 & \underline{93.67} & \underline{89.4} & \underline{56.95} & \underline{88.82} & 82.97\\
    \rowcolor{lavender!30}
    \nameframework{} (DAG)  & 110M & 1.46 & \textbf{68.9} & \textbf{84.11} & \underline{89.78} & \underline{89.77} & \textbf{77.98} & \textbf{93.69} & \textbf{90.36} & \textbf{57.08} & 88.68 & \textbf{83.93} \\
    \bottomrule
\label{tab:gmamba_glue}
\end{tabular}
} %< \resizebox
\end{table*}

\subsection{Approximate \frameworkname{} for General Topology}
\label{sec:approx_chimera_finite_sum}
While DAGs allow for efficient computation in structured domains like images and language, directly computing the resolvent \(\bL\) for general graph topology remains computationally expensive.
To address this, we use a finite-sum relaxation of the resolvent operator and truncate its corresponding Neumann series sum (Eq. \ref{eq:neumann}) at some maximum power \(k \in \sN > 0\). 
Specifically, let \(\bA\) 
be the \blue{(weighted)} adjacency matrix of the graph topology defined in Section \ref{sec:general_graphs}, then,
\begin{align}
    \bL = \sum_{i=0}^\infty \bA^i \approx \hat{\bL} = \sum_{i=0}^k \bA^i.
\end{align}
We choose \(k = \text{dia}(\gG)\), the diameter of the graph, to ensure that \(\hat{\bL}\) has access to the global structure of the graph, that is, it includes contributions from every edge and node in the graph.
\begin{proposition}
    If \(k \geq \text{dia}(\gG)\), then for any pair of nodes \((i, j)\), if \(\bL_{ij} > 0\) in the original method, then \(\hat{\bL}_{ij} > 0\) in the finite-sum relaxation.
\end{proposition}
As in Section \ref{sec:chimera_dag_efficiency}, we can compute this approximation efficiently using the squaring trick:
\begin{align}
    \hat{\bL} = (\bI + \bA)(\bI + \bA^2)(\bI + \bA^4) \cdots (\bI + \bA^p),
\end{align}
where \(p\) is the smallest power of 2 larger than or equal to the graph diameter \(\text{dia}(\gG)\). 
This reduces the computational cost of the method to $O(\log(\text{dia}(\gG)))$ matrix multiplications.

For general graphs, since we cannot exploit the underlying structure of 
adjacency matrix $A$ to design efficient algorithms, the worst-case
complexity of the finite sum approximation remains quadratic. This highlights a crucial point: as the complexity of the underlying structure increases, the computational cost of calculating the matrix $L$
will also grow accordingly.

\section{Experiments}
\label{sec:experiments}

In this section, we will demonstrate that 
\emph{directly incorporating topology 
is a powerful inductive bias 
for diverse domains} such as language, images 
and graphs, 
eliminating the need for domain-specific heuristics.
% We demonstrate that \emph{directly incorporating topology is a powerful inductive bias for disparate domains} such as language, images, and graphs, eliminating the need for domain-specific heuristics.  
\nameframework{} consistently achieves state-of-the-art performance in these domains.
On language, it outperforms BERT 
on the GLUE benchmark~\citep{glue}
by a GLUE score of 0.7. 
On images, 
it surpasses ViT models 
on the ImageNet-1k classification~\citep{imagenet}
task by 2.6\%. 
On general graphs, 
\nameframework{} outperforms strong baselines
on the Long Range Graph Benchmark~\citep{dwivedi2021graph}
which highlights our method's ability 
to model long range interactions on graphs.

\subsection{Masked Language Modeling}
We evaluate \nameframework{} on bidirectional language modeling, which has a line graph topology (Fig. \ref{fig:directed_line_graph_decomposition}). 
We test two 
\nameframework{} 
variants: the general method\footnote{
We use a slightly modified normalization scheme for the undirected line graph method to allow
for larger selectivity values in the adjacency matrix. See Appendix \ref{app:undirected_line_graphs} for details
}
(Sec. \ref{sec: sec3}) applied to an undirected line graph, and the DAG method (Sec. \ref{sec:mamba-dag}), applied to the canonical DAG decomposition of undirected line graphs into two directed line graphs and summing the resolvents of both DAGs (Fig. \ref{fig:directed_line_graph_decomposition}).
Both methods are trained on the Masked Language Modeling (MLM)~\citep{bert} task on the C4 dataset~\citep{c4} for 70k steps, following the recipe used in M2~\citep{m2}. 
The models are then fine-tuned on the GLUE benchmark. 
\blue{For baselines, we compare our methods with other sequence mixers such as M2~\cite{m2}, 
as well as models such as MLP-Mixer~\cite{mlpmixer} and FNet~\cite{fnet}, in addition to the Transformer based
BERT model~\cite{bert}. We choose these baselines to highlight the performance of our model when compared 
to different type of sequence mixing paradigms applied to large scaled language modeling tasks.}
We refer the reader to Appendix \ref{sec:Architecture_Details} for the architectural and hyper-parameter details. 

From Table~\ref{tab:gmamba_glue}, observe that while BERT outperforms other linear baselines such as M2, MLP-Mixer, FNet it does so with an additional quadratic cost. 
In contrast, \nameframework{} achieves the best of both worlds, incurring a linear time complexity while achieving strong performance. 
This capability arises from two key factors: first, our parameterization of the adjacency matrix allows the model to effectively modulate the influence between tokens in the sequence, leading to strong performance. 
\looseness=-1
Second, the structured nature of the adjacency matrix enables a fast, linear-time resolvent operation, improving the method's computational efficiency.
Additionally, note that our undirected graph (UG) variant performs competitively with BERT while surpassing other recent linear baselines. 

\subsection{ImageNet-1k Classification}

\begin{table}[h]
\centering
\begin{minipage}[t]{0.48\linewidth}
    \centering
    \caption{Top-1, Top-5 accuracies of various methods on ImageNet-1K. Chimera outperforms the standard attention baseline ViT-B, as well as other sub-quadratic baselines.}
    \resizebox{\linewidth}{!}{
        \begin{tabular}{lcccc}
            \toprule
            \multirow{2}{*}{Method (88M)} & \multicolumn{2}{c}{Top-1 (\%)} & \multicolumn{2}{c}{Top-5 (\%)}\\
            \cmidrule(lr){2-3}\cmidrule(lr){4-5}
            & Acc & Acc$_{{\text{EMA}}}$ & Acc & Acc$_{{\text{EMA}}}$\\
            \midrule
            ViT-B & 78.8 & \underline{80.6} & \underline{94.2} & \underline{95.2} \\
            S4-ViT-B & \underline{79.4} & 80.4 & \underline{94.2} & 95.1 \\
            Hyena-ViT-B & 78.4 & 76.4 & 94.0 & 93.0 \\
            \rowcolor{lavender!30} \nameframework-ViT-B & \textbf{81.4} & \textbf{82.1} & \textbf{95.4} & \textbf{95.9} \\
            \bottomrule
        \end{tabular}
    }
    \label{tab:imagenet_new}
\end{minipage}
\hfill
\begin{minipage}[t]{0.48\linewidth}
    \centering
    \caption{Ablation: Comparing 2D grid structure with 1D flattening of patches. 
    \blue{We see that maintaining the 2D DAG structure outperforms method where the underlying
    topological structure is flattened, showing maintaining the topological structure matters.}}
    \resizebox{\linewidth}{!}{
        \begin{tabular}{lcccc}
            \toprule
            \multirow{2}{*}{Method (22M)} & \multicolumn{2}{c}{Top-1 (\%)} & \multicolumn{2}{c}{Top-5 (\%)}\\
            \cmidrule(lr){2-3}\cmidrule(lr){4-5}
            & Acc & Acc$_{\text{EMA}}$ & Acc & Acc$_{\text{EMA}}$\\
            \midrule
            Fwd (1D)         & 73.8  & 73.8 & 91.6 & 91.6 \\
            Fwd \& Rev (1D)  & \underline{76.5}  & \underline{75.6} & \underline{93.4} & \underline{92.8} \\
            \rowcolor{lavender!30} 2D DAG & \textbf{77.8}  & \textbf{76.7} & \textbf{93.9} & \textbf{93.5} \\
            \bottomrule
        \end{tabular}
    }
    \label{tab:imagenet_str}
\end{minipage}
\end{table}

\vspace{-2pt}
We evaluate \nameframework{} on the ImageNet-1k~\citep{imagenet} classification task that has a grid graph topology. 
We compare \nameframework{} applied to the 2D-DAG decomposition (Figure \ref{fig:2D_grid_decomposition}) topology against
state-of-the-art ViT based models, 
a standard architecture that is used for ImageNet classification.
Specifically we use ViT-B which has 88M parameters as well as 
other SSM based baselines like Hyena~\citep{hyena}, S4~\citep{s4} in Table \ref{tab:imagenet_new}
\blue{to highlight how our method compares with other state-space model based architectures.}
We note that 
\emph{all these baselines flatten the image into a 1D sequence and apply 1D sequence models, and do not take into account the underlying topology}.
For our experiments, we simply replace the SSD layer in the Mamba block
introduced in \citet{ssd} with \nameframework{}, and use the ViT-B training recipe with minimal hyperparameter tuning.\looseness=-1

Table~\ref{tab:imagenet_new} shows that 
\nameframework{'s} 2D-DAG decomposition outperforms ViT by {2.6\%}. We note that
our method does not require any additional position embeddings
which are still an active area
of research for ViT~\citep{heo2024rotary}.
We outperform methods 
such as Hyena~\citep{hyena} by {3\%}, and S4~\citep{s4}
by {2\%} that linearize the data and then apply an SSM
on it. 

% This underscores the importance of directly incorporating topology which is a strong inductive bias.
\AGR{I think this section doesn't really spell out explicitly the main point that \emph{all other baselines flatten the image into a 1D sequence and apply 1D sequence models, and don't take into account the topology}. you probably think it's obvious to anyone who knows what the baselines are, but keep hammering the point home. you'd be surprised how many readers will miss main points when you don't spell things out.}
\AGR{I see one reference to "linearize the data" but that actually is more confusing, because you're using "linear" in multiple ways; be more consistent to use "linaer" as "linear-time" and choose a different word or phrase to talk about flattening data as a sequence}
\AGR{Similarly, the tables can be presented better in ways that highlight the differences in methods. Use tools such as section dividers to split "1D" vs "2D" methods. Be more explicit in the captions}

\begin{figure}[b]
    \centering
    \vspace*{-1em}
    \includegraphics[width=0.60\linewidth]{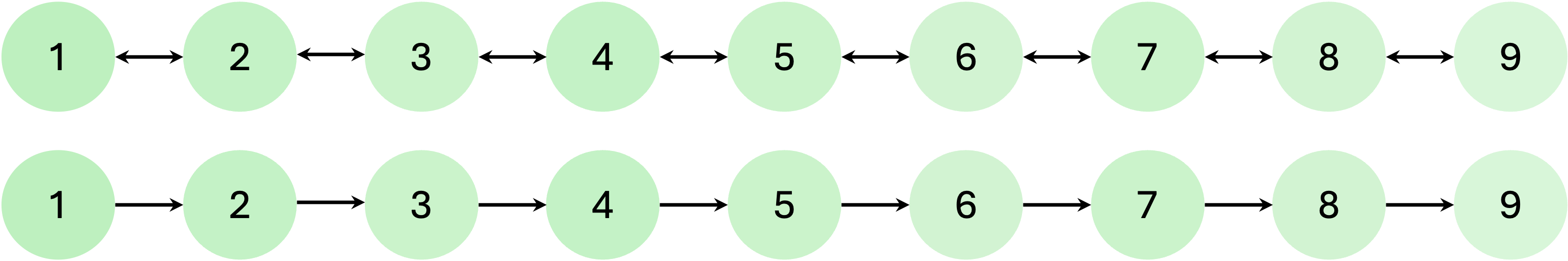}
    \caption{
        Progressively destroying the 2D grid graph topology.
        \emph{Fwd \& Rev} (top): 1D flattened grid with bidirectional edges. 
        \emph{Fwd} (bottom): 1D flattened grid graph with only forward edges.
}
    \label{fig:imagenet_ablation}
\end{figure}

Furthermore, to demonstrate the importance of incorporating topology, we perform an ablation where we progressively degrade the grid-graph structure, observing a monotonic drop in performance. 
We consider three topologies:
\textbf{2D DAG} is the 2D DAG decomposition that retains the grid structure (Fig \ref{fig:2D_grid_decomposition}, right); 
\textbf{Fwd \& Rev (1D)} flattens the grid into a 1D sequence with bidirectional edges like ViT (Fig \ref{fig:imagenet_ablation}, top);
\textbf{Fwd (1D)} is a 1D graph with only forward edges (Fig \ref{fig:imagenet_ablation}, bottom). 
We observe from Table \ref{tab:imagenet_str} that as the topology is lost, the accuracy drops from 77.8\% (2D-DAG) to 76.5\% (Fwd \& Rev) to 73.8\% (Fwd).

% We consider three topologies:
% \textbf{2D DAG} is the 2D DAG decomposition that retains the grid structure (Fig \ref{fig:2D_grid_decomposition}, right); 
% \textbf{Fwd \& Rev (1D)} flattens the grid into a 1D sequence with bidirectional edges like ViT (Fig \ref{fig:imagenet_ablation}, top);
% \textbf{Fwd (1D)} is a 1D graph with only forward edges (Fig \ref{fig:imagenet_ablation}, bottom). 
% We observe from Table \ref{tab:imagenet_str} that as the topology is lost, the accuracy drops from 77.8\% (2D-DAG) to 76.5\% (Fwd \& Rev) to 73.8\% (Fwd).

\AGR{I think these ablations are very underdescribed. I don't think a reader will be able to infer what the "fwd" and "fwd+rev" are doing at all. try putting yourself in a reader's shoes: the phrase "linearize a 2D graph into a 1D forward-backward directed graph" actually has no meaning. related, see above point about the choice of the word "linearize". At the minimum, you should have more complete descriptions of each model's architecture in the appendix and put a forward pointer to the appendix}
\AGR{in other words, you need to spell out that what you're ablating is using chimera as a 2D model, vs. applying 1D mamba (both unidirectional and bidirectional) in the same way that ViTs are applied on a flattened graph. this should be described explicitly in plain english instead of burying it in fancy terminology}

\subsection{Long Range Graph Benchmark}
\begin{table*}[t]
\caption{
Evaluation of 
\nameframework{} on LRGB Tasks~\citep{dwivedi2022long}.
The first section shows the best performing
numbers cited in the papers that introduce the
given baselines. The second section 
shows the result of better hyperparameter 
tuned baselines introduced by~\citet{tonshoff2023did}. Finally, we also
compare with other baselines that use SSMs as a 
blackbox replacement for a Transformer.
\blue{Here the best numbers are highlighted in \textbf{bold}.}}
\vspace{-2mm}
\centering
\small
\resizebox{\linewidth}{!}
{ %< auto-adjusts font size to fill line
\begin{tabular}{lccccc} %@{}}
\toprule
\multirow{2}{*}{Method (< 500k params)} & 
% \multirow{2}{*}{\#Params} & 
Peptides-Func & Peptides-Struct & PascalVOC-SP & COCO-SP\\
\cmidrule(lr){2-5}
& AP ($\uparrow)$ & MAE ($\downarrow$) & F1 ($\uparrow$) & F1 ($\uparrow)$\\
\midrule
GCN~\citep{kipf2016semi} & 0.5930$\pm$0.0023 & 0.3496$\pm$0.0013 & 0.1268$\pm$0.0060 & 0.0841$\pm$0.0010 \\
GINE~\citep{hu2019strategies} & 0.5498$\pm$0.0079 & 0.3547$\pm$0.0045 & 0.1265$\pm$0.0076 & 0.1339$\pm$0.0044 \\
Gated-GCN~\citep{bresson2017residual} & 0.5864$\pm$0.0077 & 0.3420$\pm$0.0013 & 0.2873$\pm$0.0219 & 0.2641$\pm$0.0045 \\
SAN+LapPE~\citep{kreuzer2021rethinking} & 0.6384$\pm$0.0121 & 0.2683$\pm$0.0043 & 0.3230$\pm$0.0039 & 0.2592$\pm$0.0158 \\
Exphormer~\citep{shirzad2023exphormer} & 0.6527$\pm$0.0043 & 0.2481$\pm$0.0007 & 0.3975$\pm$0.0037 & 0.3430$\pm$0.0108 \\
GPS+BigBird~\citep{rampavsek2022recipe} & 0.5854$\pm$0.0079 & 0.2842$\pm$0.0130 & 0.2762$\pm$0.0069 & 0.2622$\pm$0.0008 \\
GraphGPS+Transformer~\citep{rampavsek2022recipe} & 0.6575$\pm$0.0049 & 0.2510$\pm$0.0015 & 0.3689$\pm$0.0131 & 0.3774$\pm$0.0150 \\
\midrule
GCN~\citep{tonshoff2023did} & 
% $\leq 500$k & 
    ${0.6860} \pm 0.0050$ &  $0.2460 \pm 0.0007$ & $0.2078 \pm 0.0031$ & $0.1338 \pm 0.0007$ \\
Gated-GCN~\citep{tonshoff2023did} & 
% $\leq 500$k & 
    ${0.6765} \pm 0.0047$ &  $0.2477 \pm 0.0009$ & $0.3880 \pm 0.0040$ & $0.2922 \pm 0.0018$ \\
GINE~\citep{tonshoff2023did} & 
% $\leq 500$k & 
    $0.6621 \pm 0.0067$ &  ${0.2473} \pm 0.0017$ & $0.2718 \pm 0.0054$ & $0.2125 \pm 0.0009$ \\
GraphGPS+Transformer~\citep{tonshoff2023did} & 
% $\leq 500$k & 
    $0.6534 \pm 0.0091$ &  $0.2509 \pm 0.0014$ & ${0.4440} \pm 0.0054$ & $0.3884 \pm 0.0055$ \\
\midrule
Graph-Mamba~\citep{graphmamba} & ${0.6739} \pm 0.0087$ & $0.2478 \pm 0.0016$ & $0.4191 \pm 0.0126$ & $0.3960\pm 0.0175$ \\
Graph Mamba~\citep{graphmamba2} & $0.7071 \pm 0.0083$  & ${0.2473} \pm 0.0025$ & $0.4393 \pm 0.0112$ & ${0.3974}\pm 0.0101$ \\
\midrule
NeuralWalker~\citep{chen2024learning} & \bm{$0.7096 \pm 0.0078$} & $0.2468 \pm 0.0005$ & \bm{$0.4912 \pm 0.0042$} & \bm{$0.4398 \pm 0.0033$} \\
\midrule
\rowcolor{lavender!30} \nameframework{} (Ours) & 
    % \bm{$0.7021 \pm 0.003$}  &  \bm{$0.2433 \pm 0.0006$} & \bm{$0.4460 \pm 0.007$} & \bm{$0.3977\pm 0.016$} \\
   $0.7021 \pm 0.003$  &  \bm{$0.2433 \pm 0.0006$} & $0.4460 \pm 0.007$ & $0.3977\pm 0.016$ \\
\bottomrule
\end{tabular}
} %< \resizebox
\label{tab:graph_benchmark_results}
\end{table*}
We evaluate \nameframework{}
on the Long Range Graph Benchmark (LRGB)~\citep{dwivedi2022long}. 
This benchmark comprises tasks designed to challenge models in their ability to effectively capture both local and long-range interactions within graph structures.
% We evaluate \nameframework{} on the Long Range Graph Benchmark (LRGB) the general formulation introduced in Section \ref{sec:general_graphs}. 
We compare against convolution-based (GCN~\cite{kipf2016semi}, GatedGCN~\cite{bresson2017residual}),
Transformer-based (GraphGPS~\cite{rampavsek2022recipe}) \AGR{capitalize Transformer}, 
Mamba-based (Graph-Mamba~\cite{graphmamba}, Graph Mamba~\cite{graphmamba2}), and other baselines like GINE~\cite{hu2019strategies}, as well as 
their hyperparameter tuned versions introduced in~\citet{tonshoff2023did}.
These baselines incorporate topology using a variety of techniques: 
convolution ones use local aggregation, transformer ones use local and global aggregation via position embeddings, and Mamba ones use ``data flattening'' along with random walks, position embeddings, and local encodings.
The diversity of these methods highlights the significant research effort dedicated to heuristics to incorporate topology, in contrast to our unified approach.
% \blue{More recently, methods like NeuralWalker~\cite{chen2024learning} have achieved state-of-the-art
% result on LRGB. However, unlike Chimera and the baselines that we compre our model with, this method 
% introduces some structural modifications by sampling subgraphs via random walks and modeling them sequentially, which effectively transforms the underlying graph structure. 
% We veiw methods like NeuralWalker as complementary to our work, which could potentially be combined with Chimera layers.}

\blue{We note that while recent methods like NeuralWalker~\cite{chen2024learning} achieves state-of-the-art results on LRGB, but unlike Chimera and our chosen baselines, it modifies the graph structure by sampling subgraphs via random walks and modeling them sequentially. As our goal is to evaluate models that operate directly on the original graph. That said, NeuralWalker is complementary and could potentially be combined with Chimera layers.}

% Notably, we observe that on both localized tasks, such as Peptides-Func and Peptides-Struct, where convolution-based models typically outperform transformers, as well as on global context tasks, 
% like PascalVOC and COCO, where transformers outperform convolution models, 
% \nameframework{} consistently surpasses all baselines, with a more than 12\% improvement on PascalVOC. 
% This validates our grounded approach which effectively captures both local and global information.

We show that \nameframework{} achieves strong performance across all LRGB tasks (Table \ref{tab:graph_benchmark_results}). 
Notably, we observe that on 
tasks such as 
Peptides-Func and Peptides-Struct, 
where convolution-based models
typically outperform 
transformers, \nameframework{}
outperforms or matches their performance.
Furthermore, on tasks 
like PascalVOC 
and COCO where transformers do well, 
\nameframework{} is competitive with the best baselines. 
This validates our grounded approach which effectively captures both local and global information.

In Table \ref{tab:chimera_ablation}, we evaluate the approximate variant of \nameframework{} with a finite-sum relaxation (Sec \ref{sec:approx_chimera_finite_sum}) that truncates the Neumann series at the  diameter of the graph. 
We show that the approximation variant matches the strong transformer baseline of GraphGPS, however fully leveraging the entire graph structure in \nameframework{} provides clear performance benefits.

\begin{table*}[b]
\centering
\small
\caption{
Ablation: \nameframework{} with approximate resolvent is competitive with the Transformer baseline. This alleviates the cubic cost of evaluating the exact resolvent.
}
\resizebox{0.85\linewidth}{!}{
    \begin{tabular}{lcccc}
        \toprule
        \multirow{2}{*}{Method} & 
        Peptides-Func & Peptides-Struct & PascalVOC-SP & COCO-SP \\
        \cmidrule(lr){2-5}
        & AP ($\uparrow$) & MAE ($\downarrow$) & F1 ($\uparrow$) & F1 ($\uparrow$) \\
        \midrule
        GraphGPS+Transformer & $0.6534 \pm 0.0091$ & $0.2509 \pm 0.0014$ & $0.4440 \pm 0.0054$ & $0.3884 \pm 0.0055$ \\
        \rowcolor{pink!30} \nameframework{} (Approx) & $0.6979 \pm 0.0057$ & \bm{$0.2420 \pm 0.0013$} & $0.4353 \pm 0.00307$ & $0.3833 \pm 0.0006$ \\
        % \rowcolor{lavender!30} \nameframework{} (Ours) & $0.7021 \pm 0.003$ & $0.2433 \pm 0.0006$ & $0.446 \pm 0.007$ & $0.3977 \pm 0.016$ \\
        \rowcolor{lavender!30} \nameframework{} (Ours) & \bm{$0.7021 \pm 0.003$} & $0.2433 \pm 0.0006$ & \bm{$0.446 \pm 0.007$} & \bm{$0.3977 \pm 0.016$} \\
        \bottomrule
    \end{tabular}
    } 
\label{tab:chimera_ablation}
\vspace{-0.4cm}
\end{table*}

\section{Discussion and Future Work}

We propose \nameframework{}, a unified framework that directly incorporates the underlying graph topology in a principled way.  
Unlike prior approaches that apply attention or State Space Models (SSMs) by flattening the data, we instead generalize SSMs---originally designed to operate on sequences without position embeddings---to any graph topology.  
We show that \nameframework{} achieves strong performance across domains including language, vision, and graph tasks, consistently surpassing baselines, which validates our premise and the proposed solution.  
We further show that for the subclass of graphs which can be decomposed into DAGs, the recurrent form of \nameframework{} affords linear complexity.  

Our work has a few limitations, the most significant being that for general graphs, fully capturing all node interactions results in a cubic computational cost.  
This can be reduced to a quadratic cost through a straightforward approximation that truncates the infinite sum to a finite terms---determined by the diameter of the graph.  
That said, we still believe there is significant potential for hardware optimization just as Mamba-based methods benefited greatly from dedicated CUDA kernels.  
Nevertheless, we note that the current implementation of \nameframework{} achieves a reasonable time ratio of $\sim1.5\times$ compared to Transformer-based architectures which we believe provides a starting point for further exploration across novel domains.

We believe that developing optimized kernels for specific graph structures \blue{such as grid-graphs}---along with exploring graph approximations through DAG decompositions---is a promising direction for future work.  
We are hopeful that the community will apply \nameframework{} to a broader range of domains with inherent topological structures and continues to develop more efficient and performant extensions of \nameframework{}.

\bibliographystyle{plainnat}
\bibliography{references}

\appendix
\onecolumn

\section{Deferred Proofs}
\subsection{Proof of Proposition \ref{prop:row_wise_normalization}}
\begin{proof}
    \label{proof:row_wise_normalization}
    Let \blue{\(\boldsymbol{\epsilon}_i \sim \mathcal{N}(\mathbf{0}, \mathbf{I}_T)\)} be \(T\) i.i.d. random Gaussian vectors. Assuming Gaussian initialization for the adjacency matrix \(\mathbf{A}\), it can be expressed as:
    \begin{align}
        \mathbf{A}[i,:] = \frac{
            \gamma \boldsymbol{\epsilon}_i
        }{
            \|\boldsymbol{\epsilon}_i\| + \exp(-\Psi_i)
        }.
    \end{align}
    We first show that $\|\bA\| \le \gamma < 1$.
    From the concentration of the Gaussian random vector norm,  \blue{$\|\boldsymbol{\epsilon}_i\| \ge \sqrt{T}/2$ for all tokens $i$, with probability $\ge 1 - \exp(T/8)$. 
    Since $\exp(-\Psi_i) \ge 0$, 
    $\|\boldsymbol{\epsilon}_i\|+\exp(-\Psi_i) \ge \sqrt{T}/2$}.
    Consider any unit vector $\vu$, then 
    \begin{align}
        \|\bA \vu\| 
        = 
        \sum_{i=1}^T
        \frac{
            \gamma \boldsymbol{\epsilon}_i^T\vu
        }{
            \|\boldsymbol{\epsilon}_i\| + \exp(-\Psi_i)
        }
        \le
        \blue{
        \gamma 
        \sum_{i=1}^T
        \frac{
            2\epsilon_i
        }{
            \sqrt{T}
        }
        \le 
        2\gamma 
        \frac{
            \sqrt{T}\epsilon
        }{
            \sqrt{T}
        } = 2\gamma \epsilon < 1,}
    \end{align}
    with probability greater than \blue{$1-\Phi(\tfrac{-1}{2\gamma})$}, were $\epsilon_i, \epsilon \sim \gN(0,1)$.
    Finally, since the operator norm of $\|\bA\|$ is less than one, we apply Banach's Lemma to get,
    \begin{align}
        \|(\bI - \bA)^{-1} \| \leq \frac{1}{1 - \|\bA\|},
    \end{align}
    which implies that the inverse exists.
\end{proof}

\subsection{Proof of Proposition \ref{prop:dag_normalizer}}
\begin{proof}
\label{proof:dag_normalizer}
    \begin{align}
        \text{Var}(\bC_i^T\vh_i) 
        &= 
        \frac{1}{|p(i)|}
        \left(
        \sum_{j \in p(i)} 
            \bA_{ij} \text{Var}(\bC_i^T\vh_j) + \ln(\bA_{ij}) \text{Var}(\bC_i^T\bB_i v_i)
        \right),
        \\
         &=
        \frac{1}{|p(i)|}
        \left(
        \sum_{j \in p(i)} 
            \bA_{ij} \text{Var}(\bC_j^T\vh_j) + \frac{2}{d}\ln(\bA_{ij})
        \right),
    \end{align}
    where we have used the fact that $\text{Var}(\bC_j^T\vh_j) = \text{Var}(\bC_i^T\vh_j)$, and that the variance of $\gX^2$ distribution with $d$ degrees of freedom is $2d$. Let $d \ge 4$, then
    \begin{align}
        \text{Var}(\bC_i^T\vh_i) 
        &\le 
        \frac{1}{|p(i)|}
        \left(
        \sum_{j \in p(i)} 
            \bA_{ij} + \frac{2}{d}\ln(\bA_{ij})
        \right)
        \le 
        \frac{1}{|p(i)|}
        \sum_{j \in p(i)}  1 \le 1,
    \end{align}
    where we have used the fact that $\bA_{ij} \in [0,1]$.
\end{proof}

% \subsection{Proof of Proposition \ref{prop:linear}}
% \begin{proof}
% \label{proof:linear}
%     In the structured masked attention (SMA) framework~\cite{ssd}, the computational complexity is the cost of the matrix-vector multiplication by the mask matrix $\bL = (\bI - \bA)^{-1}$.
%     % \AG{Make sure that we state this fact in the preliminaries, perhaps as a proposition}.
%     In the case of DAGs, $\bA$ is (up to conjugation by a permutation) a \emph{lower-triangular} matrix with $|\gE|$ (number of edges) non-zero entries.
%     It suffices to analyze the cost of computing the multiplication $\vy = (\bI - \bA)^{-1} \vx$.
%     Rewriting as $(\bI - \bA) \vy = \vx$, $\vy$ can be computed through Gaussian elimination on the matrix $\bI - \bA$,
%     which takes time proportional to the number of non-zero entries or $|\gV|+|\gE|$.
    
%     In graph terminology, this operation can be viewed as a dynamic programming algorithm to propagate features through the SSM update, where the ordering of edges to perform the update rule is given by the Gaussian elimination ordering.
% \end{proof}

\subsection{Proof of Proposition \ref{prop:linear}}  
\begin{proof}  
\label{proof:linear}  
    In the structured masked attention (SMA) framework~\cite{ssd}, the computational complexity is the cost of the matrix-vector multiplication by the mask matrix $\bL = (\bI - \bA)^{-1}$.
    For DAGs, $\bA$ is (up to conjugation by a permutation) a \emph{lower-triangular} matrix with $|\gE|$ nonzero entries. Computing $\vy = (\bI - \bA)^{-1} \vx$ reduces to solving the system $(\bI - \bA) \vy = \vx$ via forward substitution.
    
    We perform Gaussian elimination by iterating over the ordered list $\{0, \dots, |\gV|-1\}$ and choosing the pivots $(i,i)$. 
    Since $\bI - \bA$ is lower-triangular, each pivot operation affects only a single column rather than the entire row, reducing the cost per step to $O(\text{nnz}(\mathbf{A}[:, i]))$, where $\text{nnz}(\cdot)$ denotes the number of non-zero entries.
    Summing over columns, the complexity is,
    \[
    O\left(\sum_{i}^{|\gV|} \text{nnz}(\mathbf{A}[:, i])\right) = O(\text{nnz}(\mathbf{A})) = O(|\gV| + |\gE|).
    \]
    For our motivating example of Mamba, $\bI - \bA$ has exactly $2|\gV|$ nonzero entries, ensuring a linear-time complexity.
\end{proof}

\subsection{Proof of Theorem \ref{thm:dagthm}}
\begin{proof}
    \label{proof:dagthm}
    
    \textbf{Backward pass.  }
    The local update rule of backpropagation requires applying the chain rule through the matrix inverse operation,
    in particular, using the following identity applied to $\bY = (\bI - \bA)$,
    \begin{align}
        \frac{\partial \bY^{-1}}{\partial \theta} =
        -\bY^{-1}
        \frac{\partial \bY}{\partial \theta}
        \bY^{-1}
    \end{align}
    Because $\bY^{-1}$ is already computed in the forward pass, it can be cached,
    and then the marginal cost of the local backpropagation is simply two extra matrix multiplications.

    \textbf{Forward pass.  }
    To compute $\bL = (\bI - \bA)^{-1}$ more efficiently for DAGs, we leverage the equivalence of Neumann series to the series $\bL = \bI + \bA + \bA^2 + \cdots$, which comes to a finite sum for DAGs due to the nilpotence of $\bA$ matrix.
    We compute this sum more efficiently using the ``squaring trick'' as, 
    \begin{align}
        (\bI - \bA)^{-1} = (\bI + \bA)(\bI + \bA^2)(\bI + \bA^4) \cdots (\bI + \bA^k),
    \end{align}
    where $k$ is the smallest power of 2 larger than the graph diameter $\text{dia}(\gG)$.
    This can be computed using $O(\log(\text{dia}(\gG)))$ matrix multiplications to compute the powers of $\bA$ for powers-of-two exponents,
    and then $O(\log(\text{dia}(\gG)))$ matrix multiplications to multiply together the right-hand side.
\end{proof}
\newpage
\section{Additional Experiments}

\subsection{MLM: \nameframework{} on Undirected Line Graphs}
\label{app:undirected_line_graphs}
For an undirected line graph (Figure~\ref{fig:directed_line_graph_decomposition}, left), the adjacency matrix $\bA$ takes the following form:
\begin{align*}
\bA = \begin{bmatrix}
0 & a_{12} & 0 & \cdots & 0 \\
a_{21} & 0 & a_{23} & \cdots & 0 \\
0 & a_{32} & 0 & \cdots & 0 \\
\vdots & \vdots & \vdots & \ddots & \vdots \\
0 \cdots 0 & 0 \cdots 0 & 0 & a_{T-1, T} & 0
\end{bmatrix}.
\end{align*}
As discussed in Section~\ref{sec:general_graphs}, to ensure the existence of $(\bI - \bA)^{-1}$, we introduced a row-wise sum normalization strategy, wherein we normalized each row of the adjacency matrix with $\sum_j \bA_{ij} + \Psi_i$. 
However, since this constraint is designed for general graphs, it is not sufficiently expressive. Therefore, we instead use a strictly more expressive constraint for line graphs which enforces $\bA_{ij} \cdot \bA_{ji} + \Psi_i \leq \frac{1}{4}$ on each simple cycle of the graph. 

\begin{proposition}
    Under the above constraint, the inverse $(\bI - \bA)^{-1}$ exists as for any two nodes, the sum of all paths between them is upper bounded by $\sum_i (1/4)^i \le 1/3$.
\end{proposition}

\subsection{Imagenet: Parameter Sharing Ablation}
\label{app:additional_experiments}

We study the trade-off between sharing parameters for $\bB, \bC$ across different graphs as a domain-dependent design choice. 
We explore four settings: \textit{No sharing, Complete sharing, Row-wise sharing, and Diagonal sharing} across the four DAGs. 
From Table \ref{tab:imagenet_str_copy}, we observe that diagonal sharing achieves the best performance, indicating it strikes the optimal tradeoff between parameter sharing and other modes of increasing expressivity for modeling image data.

\begin{table}[H]
\centering
\begin{minipage}{0.50\textwidth}
\vspace{-0.3cm}
\centering
\resizebox{0.85\linewidth}{!}{
\begin{tabular}{lllll}
\toprule
\multirow{2}{*}{Method (22M)}         & \multicolumn{2}{c}{Top-1 (\%)} & \multicolumn{2}{c}{Top-5 (\%)}\\
\cmidrule(lr){2-3}\cmidrule(lr){4-5}
                                & Acc   & Acc$_{\text{EMA}}$ & Acc  & Acc$_{\text{EMA}}$\\
\midrule
None          & 77.10  & 76.13 & 93.55 & 93.15 \\
Complete       & 77.25  & 76.09 & 93.75 & 93.21 \\
Row-wise    & {77.46}  & {76.57} & {93.76} & {93.37} \\
\rowcolor{lavender!30} Diagonal & \textbf{77.80}  & \textbf{76.69} & \textbf{93.87} & \textbf{93.53} \\
\bottomrule
\end{tabular}
}
\caption{
Ablation: Diagonal parameter sharing works best.
}
\label{tab:imagenet_str_copy}
\end{minipage}
\end{table}

\subsection{Chimera When Both \# Layers and \# Parameters Are Controlled}
\label{app:layer_control}

Chimera's architecture builds on the Mamba Block from Mamba-2, which utilizes a greater number of layers than Transformers due to its higher parameter efficiency. In this section, we conduct an ablation study where both the number of layers and total parameter count are controlled by adjusting the expansion factor \( e \) in the Mamba Block to 4. Specifically, we compare three models on the bidirectional language modeling task:

\begin{itemize}
    \item \textbf{Chimera-12L}: 12 layers, baseline configuration with 70M parameters.
    \item \textbf{BERT-6L}: 6-layer Transformer baseline with 70M parameters.
    \item \textbf{Chimera-6L}: 6 layers with an expansion factor of 4, maintaining the same parameter count as the other models.
\end{itemize}
To reduce computational costs, we train these models on a reduced ablation setting with 98M steps instead of the standard 245M steps and the results are summarized in the table below. 
Notably, Chimera-6L and Chimera-12L achieve nearly identical performance, both significantly outperforming BERT-6L. This demonstrates that Chimera's improvements are not simply a result of increased depth but rather stem from its core methodological advancements.

\begin{table}[h]
    \centering
    \caption{Ablation: Chimera maintains strong performance when both number of layers and number of parameters are controlled.}
    \resizebox{0.50\linewidth}{!}{
        \begin{tabular}{lcc}
            \toprule
            \multirow{2}{*}{Model} & \multicolumn{2}{c}{Evaluation Metrics} \\
            \cmidrule(lr){2-3}
            & Masked Accuracy ($\uparrow$) & Cross-Entropy Loss ($\downarrow$) \\
            \midrule
            BERT-6L & 0.6176 & 1.9466 \\
            \rowcolor{pink!30} Chimera-6L & \textbf{0.6360} & \textbf{1.8108} \\
            \rowcolor{lavender!30} Chimera-12L & \textbf{0.6363} & \textbf{1.8142} \\
            \bottomrule
        \end{tabular}
    }
\label{tab:chimera_layer_control}
\vspace{0.3cm}
\end{table}
\vspace{10cm}

\newpage
% \input{fig/hydra}
% \newpage

\section{Architectural Details}
\label{sec:Architecture_Details}

\begin{figure}[H]  % Use H to force the figure to appear here
    \centering
    \includegraphics[width=0.75\textwidth]{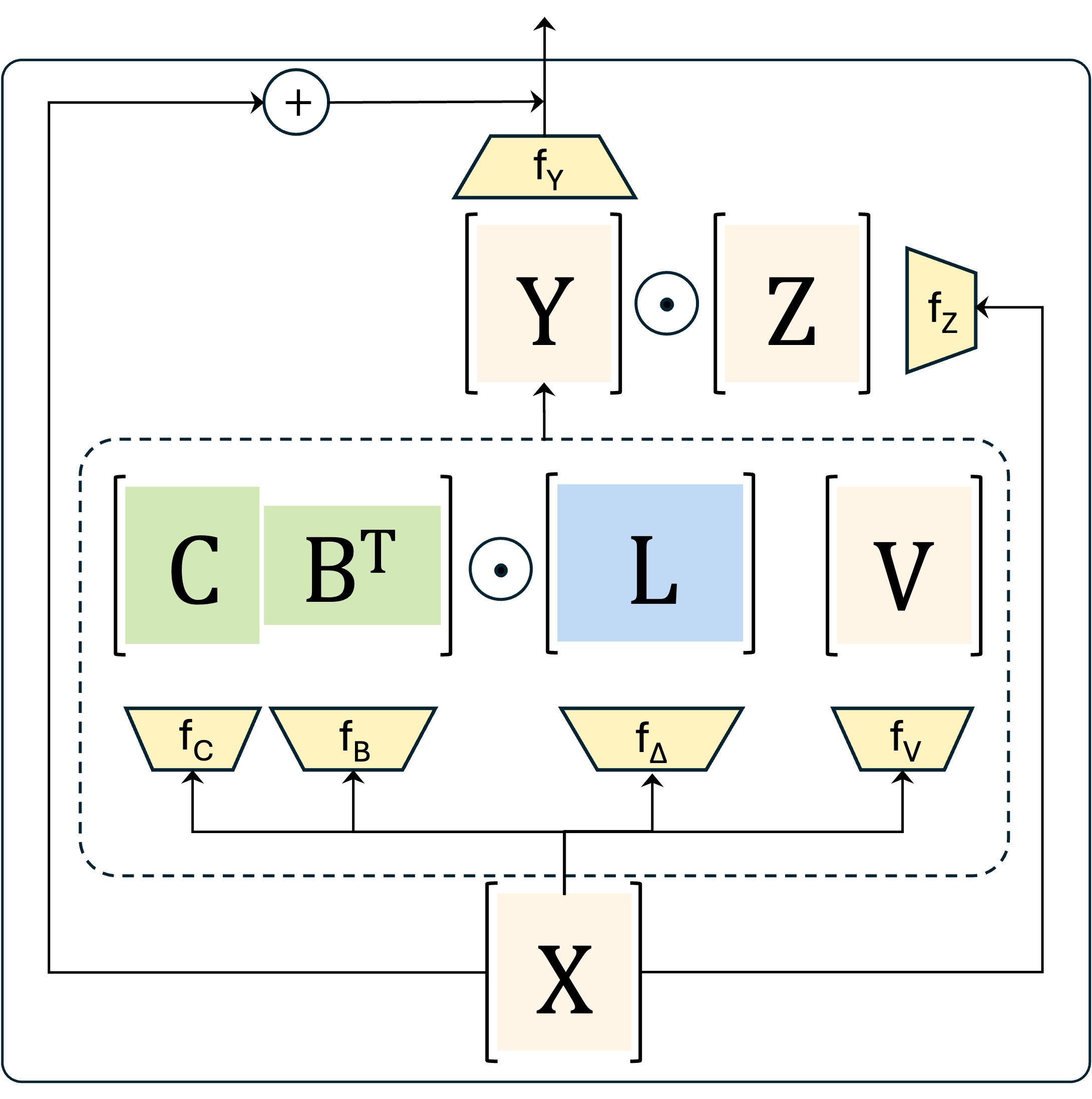}  % Adjust width as needed
    \caption{\nameframework{'s} Architecture: The output of the \nameframework{} layer is embedded within the gated block introduced in Mamba-2~\citep{ssd}. 
    Here $\bX$ matrix denotes the input to the block, and $f_c, f_B, f_{\Delta}$ and $f_V$ are data dependent projections defined in Section~\ref{sec:prelims}.
    The operator $\odot$ denotes 
    element-wise multiplications between matrices, and $\oplus$ defines addition.
    The output from the Chimera layer is passed through a Gated-MLP, a final projection $f_Y$, followed by a residual connection.}
    \label{fig:chimera_figure}
\end{figure}

\subsection{Masked Language Modeling}
\begin{table}[ht]
\centering
\caption{Architectural and Training Details for BERT-B and \nameframework{} on MLM}
\begin{tabular}{lcc}
\toprule
\textbf{Parameter} & \textbf{BERT-B} (110M) & \textbf{\nameframework{}} (110M) \\
\midrule
Model dimension ($d_{\text{model}}$) & 768 & 768 \\
\rowcolor{pink!30} Layers & 12 & 23 \\
Max sequence length & 128 & 128 \\
Num Heads & 12 & 12 \\
Head size & 64 & 64 \\
Optimizer & Decoupled AdamW & Decoupled AdamW \\
\rowcolor{pink!30}  Learning rate & $5e-4$ & $8e-4$ \\
Optimizer momentum & $\beta_1 = 0.9, \beta_2 = 0.98$ & $\beta_1 = 0.9, \beta_2 = 0.98$ \\
Weight decay & $1e-5$ & $1e-5$ \\
Batch size & 4096 & 4096 \\
Learning rate schedule & Linear decay with warmup & Linear decay with warmup \\
Training steps & 70k & 70k \\
MLM Probability & 0.3 & 0.3 \\
\bottomrule
\end{tabular}
\label{tab:mlm_hyperparameters}
\end{table}

In Table~\ref{tab:mlm_hyperparameters}, we provide the architectural and training details for BERT-B and \frameworkname{} on the MLM task. 
For both the models, we follow the M2 recipe from \citet{m2}, adjusting the number of layers to 12 for BERT-B and 23 for \frameworkname{} to control for the number of parameters.
We conducted a small sweep to fine-tune the learning rate for Chimera, choosing $8e-4$ over BERT-B's $5e-4$.

\subsection{Imagenet-1k Classification}

For the image classification experiments, we largely follow the ViT-B recipe with the following adjustments as shown in Table \ref{tab:image_hyperparameters}, \blue{where the hyperparameters are carefully tuned---and hence different from Table~\ref{tab:mlm_hyperparameters}---in order to perform best on the image classification task.}
To control for the number of parameters, we adjust the number of layers from 12 for ViT-B to 22 for \frameworkname{}. 
Additionally, we reduce the Cutmix augmentation from 1.0 to 0.1, as \nameframework{'s} stronger inductive bias mitigates the risk of overfitting. 

In Table \ref{tab:chimera_ablation_reduced}, we present the reduced setting used for our ablation studies in Tables~\ref{tab:imagenet_str_copy} and \ref{tab:imagenet_str}, where we match the number of parameters of ViT-S (22M).

\begin{table}[ht]
\centering
\caption{Hyperparameters used for ViT-B and \nameframework{} for ImageNet-1k classification task}
\begin{tabular}{lcc}
\hline
\textbf{Parameter} & \textbf{ViT-B} (88M) & \textbf{\nameframework{} (88M)} \\
\hline
Image size & $224^2$  & $224^2$\\
Optimizer & AdamW  & AdamW\\
Optimizer momentum & $\beta_1, \beta_2 = 0.9, 0.999$ &   $\beta_1, \beta_2 = 0.9, 0.999$\\
Weight init & trunc. normal (std=0.02) & trunc. normal (std=0.02) \\
Learning rate & $1e-3$ & $1e-3$\\
Weight decay & 0.05 & 0.05\\
Batch size & 1024 & 1024\\
Training epochs & 310 & 310\\
Learning rate schedule & cosine decay & cosine decay\\
Warmup epochs & 10 & 10\\
Warmup schedule & linear & linear \\
Patch Size & 16 & 16\\
\rowcolor{pink!30}  Layers & 12 & 22\\
Num Heads & 12 & 12\\
Droppath & 0.3 & 0.3\\
\hline
Randaugment & (9,0.5,layers=2) & (9,0.5,layers=2) \\
Mixup & 0.8 & 0.8\\
\rowcolor{pink!30} Cutmix & 1.0 & 0.1\\
Random erasing & 0.25 & 0.25 \\
\rowcolor{pink!30} Label smoothing & 0.1 & 0.25\\
\rowcolor{pink!30} Stochastic depth & 0.1 & 0.25 \\
Exp. mov. avg (EMA) & 0.99996 & 0.99996 \\
\hline
\label{tab:image_hyperparameters}
\end{tabular}
\end{table}

% \begin{table}[h!]
% \centering
% \caption{Hyperparameters on Peptides-Struct.}
% \begin{tabular}{lccccc}
% \toprule
%  & \textbf{GCN} & \textbf{GINE} & \textbf{GatedGCN} & \textbf{GPS} & \textbf{\nameframework}\\
% \midrule
% lr & 0.001 & 0.001 & 0.001 & 0.001 & 0.001 \\
% dropout & 0.1 & 0.1 & 0.1 & 0.1 & 0.1\\
% \#layers & 6 & 10 & 8 & 8 & 2 \\
% hidden dim. & 235 & 145 & 100 & 64 \\
% head depth & 3 & 3 & 3 & 2 \\
% PE/SE & LapPE & LapPE & LapPE & LapPE \\
% batch size & 200 & 200 & 200 & 200 \\
% \#epochs & 250 & 250 & 250 & 250 \\
% norm & - & - & - & BatchNorm \\
% MPNN & - & - & - & GatedGCN \\
% \#Param. & 488k & 492k & 445k & 452k \\
% \bottomrule
% \end{tabular}
% \end{table}

\begin{table}[ht]
\centering
\caption{Key differences between the original and the ablation setting for \nameframework{}}
\begin{tabular}{lcc}
\toprule
\textbf{Parameter} & \textbf{\nameframework{}-S (2D)} \\
\midrule
Model dimension ($d_{\text{model}}$) & 384 \\
Number of layers & 22 \\
Number of Heads & 3 \\
Droppath & 0.1 \\
\bottomrule
\end{tabular}
\label{tab:chimera_ablation_reduced}
\end{table}

\subsection{Long Range Graph Benchmark}
To train \nameframework{} on the Long Range Graph Benchmark we follow a similar training recipe to that provided in~\citet{rampavsek2022recipe}
where we replace the Transformer layers with \nameframework{} layers. 
Moreover, in line with the baselines, we make sure that 
our models have less than $500k$ parameters.
While training \nameframework{} 
on graphs 
% Keeping in line with the architectures defined in ~\citet{rampavsek2022recipe}, 
we remove
the Gated-MLP layer $Z$ defined in Figure~\ref{fig:chimera_figure}.
We did this to keep our training recipe as close to that provided in ~\citet{rampavsek2022recipe}
and highlight the effectiveness of \nameframework{}.
The hyperparameters used to train \nameframework{} are provided in Table~\ref{tab:graph_hyperparameters}.
\begin{table}[H]
\centering
\caption{Hyperparameters running \nameframework{} on the Long Range Graph Benchmark}
\begin{tabular}{lcccc}
\toprule
 & \textbf{Peptides-Func} & \textbf{Peptides-Struct} & \textbf{PascalVOC-SP} & \textbf{COCO-SP} \\
\midrule
Learning Rate & 0.001 & 0.0015 & 0.0035 & 0.0035 \\
Optimizer & Adam & Adam & Adam & Adam \\
dropout & 0.1 & 0.1 & 0.05 & 0.05  \\
\#layers & 8 & 8 & 8 & 8 \\
hidden dim. & 64 & 64 & 64 & 64 \\
hidden state dim. & 96 & 80 & 64 & 64 \\
num heads & 2 & 4 & 4 & 4 \\
batch size & 64 & 64 & 32 & 32 \\
\#epochs & 200 & 200 & 200 & 200 \\
norm & LayerNorm & LayerNorm & LayerNorm& LayerNorm \\
MPNN & GCN &  GCN & GCN & GCN\\
\#Param. & 499k & 504k & 489k &  489k \\
\bottomrule
\label{tab:graph_hyperparameters}
\end{tabular}
\end{table}

% \input{supplementary/additional_experiments}
% \input{supplementary/broader_impact}

% \section{Standard Deviations}
% \input{supplementary/tab/glue_std}

\end{document}